\documentclass[journal,11pt,onecolumn]{IEEEtran}
\usepackage{setspace}
\usepackage{cite}

\ifCLASSINFOpdf
\else
\fi

\usepackage{amsmath,amsthm,amsfonts,amscd} 
\usepackage{eucal,color} 	 	
\usepackage{verbatim}      	
\usepackage{makeidx}       	
\usepackage{psfig}         	
\usepackage{epsfig}         	

\usepackage{url}		
\usepackage{comment} 
\usepackage{graphicx,epsfig,enumitem}	
\usepackage{epstopdf}

\def\x{{\mathbf x}}

\def\y{{\mathbf y}}

\def\R{{\mathbf R}}

\def\U{{\mathbf U}}

\def\e{{\eta}}
\def\s{\star}
\def\dist{\text{dist}}

\usepackage{pbox}
\usepackage{subcaption}
\usepackage{caption}
\usepackage{tabulary}
\usepackage{algcompatible,algorithm}
\usepackage[font={footnotesize}]{caption}
\usepackage{multicol,lipsum,stfloats}

\newtheorem{property}{\textbf{Property}}

\def\s{\star}
\def\p{\prime}

\def\M{{\bf M}}
\def\R{{\bf R}}
\def\u{{\bf u}}
\def\U{{\bf U}}
\def\V{{\bf V}}
\def\w{{\bf w}}

\def\F{{\bf F}}
\def\B{{\bf B}}
\def\C{{\bf C}}
\def\S{{\bf S}}
\def\N{{\bf N}}
\def\v{{\bf v}}
\def\x{{\bf x}}

\def\y{{\bf y}}

\def\e{{\bf e}}
\def\dist{\text{dist}}
\newcolumntype{L}[1]{>{\raggedright\let\newline\\\arraybackslash\hspace{0pt}}m{#1}}
\newcolumntype{C}[1]{>{\centering\let\newline\\\arraybackslash\hspace{0pt}}m{#1}}
\newcolumntype{R}[1]{>{\raggedleft\let\newline\\\arraybackslash\hspace{0pt}}m{#1}}

\newtheorem{thm}{Theorem}
\newtheorem{cor}{Corollary}
\newtheorem{lem}{Lemma}

\theoremstyle{definition}
\newtheorem{defn}{Definition}[section]

\theoremstyle{remark}

\begin{document}

\setlength{\abovedisplayskip}{3pt}
\setlength{\belowdisplayskip}{3pt}
%
\title{Matrix Completion and Performance Guarantees for Single Individual Haplotyping}
%
%
%

\author{Somsubhra Barik and Haris Vikalo,~\IEEEmembership{Senior Member,~IEEE}
\thanks{S. Barik and H. Vikalo are with the Department
of Electrical and Computer Engineering, The University of Texas at Austin, Austin, 
TX, 78712 USA e-mail: sbarik@utexas.edu.}
}

%
%

\markboth{Matrix Completion and Performance Guarantees for Single Individual Haplotyping}%
{Matrix Completion and Performance Guarantees for Single Individual Haplotyping}
%



\maketitle

\begin{abstract}

Single individual haplotyping is an NP-hard problem that emerges when
attempting to reconstruct an organism's inherited genetic variations using data typically 
generated by high-throughput DNA sequencing platforms. Genomes of diploid organisms, 
including humans, are organized into homologous pairs of chromosomes that differ from
each other in a relatively small number of variant positions. Haplotypes are ordered 
sequences of the nucleotides in the variant positions of the chromosomes in a homologous 
pair; for diploids, haplotypes associated with a pair of chromosomes may be 
conveniently represented by means of complementary binary sequences. In this paper, we 
consider a binary matrix factorization formulation of the single individual haplotyping problem 
and efficiently solve it by means of alternating minimization. We analyze the convergence 
properties of the alternating minimization algorithm and establish theoretical bounds for the
achievable haplotype reconstruction error. The proposed technique is shown to outperform 
existing methods when applied to synthetic as well as real-world Fosmid-based HapMap 
NA12878 datasets.
\end{abstract}


%
\IEEEpeerreviewmaketitle

\section{Introduction}\label{sec:haplo_intro}
%
%
%
%

\IEEEPARstart{D}{NA} of diploid organisms, including humans, is organized into pairs of homologous chromosomes. The two
chromosomes in a pair differ from each other due to point mutations, i.e., they contain so-called single nucleotide 
polymorphisms (SNPs) in a fraction of locations. 
SNPs are relatively rare; for humans, the SNP rate between two homologous chromosomes is roughly $1$ in $300$ 
base-pairs \cite{sachidanandam}. The ordered sequence of SNPs located on a chromosome in a homologous pair is 
referred to as a {\it haplotype}. Haplotype information is of critical importance for personalized medical applications, including 
the discovery of an individual's susceptibility to diseases \cite{clark}, whole genome association studies \cite{gibbs}, gene 
detection under positive selection and discovery of recombination patterns \cite{sabeti}. High-throughput DNA sequencing 
platforms rely on so-called shotgun sequencing strategy to randomly oversample the pairs of chromosomes and generate 
a library of overlapping paired-end reads (fragments). Parts of the reads that do not cover variant positions are typically
discarded; the remaining data is conveniently organized in a read-fragment matrix where the rows correspond to reads and
columns correspond to SNPs. Since the SNPs are relatively rare and reads are relatively short, the read-fragment matrix is
typically very sparse. If the reads were free of sequencing errors, haplotype assembly would be straightforward and would 
require partitioning the reads into two clusters, one for each chromosome in a pair. However, presence of sequencing 
errors (of the order $10^{-3}$ to $10^{-2}$) gives rise to ambiguities about the origin of reads and renders the single individual
haplotyping (SIH) problem computationally very challenging. 


\begin{figure}[!t]        
\includegraphics[width=\linewidth]{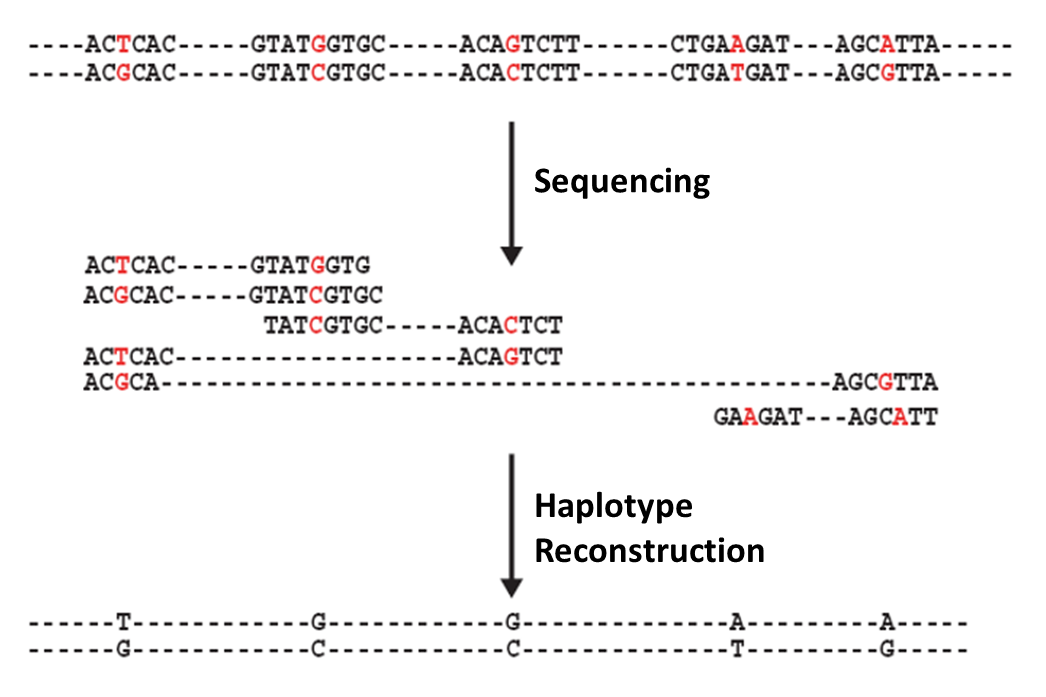}
\vspace{-0.2in}
\caption{An illustration of the single individual haplotyping problem (often also referred to as haplotype assembly). 
The nucleotides marked in red denote the SNPs on a pair of chromosomes. During sequencing, paired-end reads 
(i.e., DNA fragments) containing the SNPs are generated from multiple copies of the chromosomal sequences; one 
can think of each read as being obtained by sampling (with replacement) one of the chromosomes. We assume
that the relative ordering of reads can be determined by mapping them to a reference genome. The goal of
single individual haplotyping is to determine the order of SNPs associated with each chromosome in the pair.}
\label{fig:hap_assembly}
\end{figure}


Approaches to SIH attempt to perform optimization of various criteria including minimum fragment removal, minimum SNP 
removal and most widely used minimum error correction (MEC) objectives \cite{lancia}. Finding the optimal MEC solution 
to the SIH problem is known to be NP-hard \cite{lancia,cilibrasi}. A branch-and-bound approach in \cite{wang} solves the 
problem optimally but the complexity of the scheme grows exponentially with the haplotype length. Similar approach was 
adopted in \cite{das_sphere} where statistical information about sequencing errors was exploited to solve the MEC problem 
using sphere decoding. However, the complexity of this scheme grows exponentially with haplotype length and quickly
becomes prohibitive. Suboptimal yet efficient methods for SIH include greedy approach \cite{levy}, max-cut based solution 
\cite{bansal_hapcut}, Bayesian methods based on MCMC \cite{bansal_mcmc}, greedy cut based \cite{duitama} and 
flow-graph based approaches \cite{aguiar}.
More recent heuristic haplotype assembly approaches include a convex optimization program for minimizing the MEC 
score \cite{sdhap}, a communication-theoretic approach solved using belief propagation \cite{zrinka_TCBB}, dynamic 
programming based approach using  graphical models \cite{probhap} and probabilistic mixture model based approach 
\cite{mixsih}. Generally, these heuristic methods come without performance guarantees.

Motivated by the recent developments in the research on matrix completion (overviewed in 
Section~\ref{sec:background_matrix_compl}), in this paper we formulate SIH as a rank-one matrix 
completion problem and propose a binary-constrained variant of alternating minimization algorithm 
to solve it. We analyze the performance and convergence properties of the proposed algorithm, 
and provide theoretical guarantees for haplotype reconstruction expressed in the form of an upper 
bound on the MEC score. Furthermore, we determine the sample complexity (essentially, sequencing 
coverage) that is sufficient for the algorithm to converge. Experiments performed on both synthetic 
and HapMap sample NA12878 datasets demonstrate the superiority of the proposed framework over 
competing methods. Note that a matrix factorization framework was previously leveraged to solve
SIH via gradient descent in \cite{cai_vikalo}; however, \cite{cai_vikalo} does not provide theoretical 
performance guarantees that are established for the alternating minimization algorithm proposed in 
the current manuscript. An early preliminary version of our work was presented in \cite{icassp2017}. 

\subsection{Notation}\label{sec:notation}
Matrices are represented by uppercase bold letters and vectors by lowercase bold letters. For a matrix 
$\M$, $\M^{(i)}$ and $\M_i$ represent its $i^{th}$ row and $i^{th}$ column, respectively. $M_{ij}$ denotes 
the $(i,j)^{th}$ entry of matrix $\M$ and $u_i$ denotes the $i^{th}$ entry of vector $\u$.  $\M^{\dagger}, \|\M\|_2$, 
$\|\M\|_F$ and $\|\M\|_1$ represent respectively the transpose, the spectral norm (or 2-norm), the Frobenius 
norm and entry-wise $\ell_1$ norm (i.e., $\sum_{ij}|M_{ij}|$) of the matrix $\M$, whereas the 2-norm of a 
vector $\u\in\mathbb{R}^m$ is denoted by $\|\u\|_2 = \left(\sum_{i=1}^m|u_i|^2\right)^{1/2}$.  Each vector is 
assumed to be a column vector unless otherwise specified. A range of integers from $1$ to $m$ is denoted by 
$[m]=\{1,2,\ldots,m\}$. $\mathbb{I}$ stands for the identity matrix of an appropriate dimension. Sign of an entry 
$u_i$ is $\text{sign}(u_i)=1$ if $u_i\geq 0$,  $-1$ otherwise, and $\text{sign}(\u)$ is the vector of 
entry-wise signs of $\u$. A standard basis vector with $1$ in the $i^{th}$ entry and $0$ everywhere else is 
denoted by $\e_i$. The singular value decomposition (SVD) of a matrix $\M\in\mathbb{R}^{m \times n}$ of 
rank $k$ is given by $\M= \U\Sigma\V^T$, where $\U\in\mathbb{R}^{m \times k}$ and 
$\V\in\mathbb{R}^{n \times k}$ are matrices of the left and right singular vectors, respectively, of $\M$, with 
$\U^T \U = \mathbb{I}$ and $\V^T \V = \mathbb{I}$, and $\Sigma\in\mathbb{R}^{k\times k}$ is a diagonal 
matrix whose entries are $\{\sigma_1,\sigma_2,\ldots,\sigma_k\}$, where 
$\sigma_1\geq \sigma_2 \geq \cdots \geq \sigma_k \geq 0 $ are the singular values of $\M$. Projection of a 
matrix $\M$ on the subspace spanned by the columns of another matrix $\U$ is denoted by 
$\mathbb{P}_{\U}(\M) = \|\U\U^{T}\M\|_2$ and the projection to the orthogonal subspace is denoted by 
$\mathbb{P}_{\U^{\perp}}(\M) = \|(\mathbb{I}-\U\U^{T})\M\|_2$. Subspace spanned by vectors $\u_i$ is 
denoted by span$\{\u_i\}$.  Lastly, ${\bf 1}_{\mathcal{A}}$ denotes the indicator function for the event 
$\mathcal{A}$, i.e.,${\bf 1}_{\mathcal{A}}$ = 1, if $\mathcal{A}$ is true, 0 otherwise.  

\section{Matrix Completion Formulation of Single Individual Haplotyping}

\subsection{Brief background on matrix completion}\label{sec:background_matrix_compl}
Matrix completion is concerned with finding a low rank approximation to a partially observed matrix, and has been an active area of research in recent years. Finding a rank-$k$ approximation $\M \in \mathbb{R}^{m \times n}$, $k < \min\{m,n\}$, to a partially 
observed matrix is often reduced to the search for factors $\U\in\mathbb{R}^{m\times k}$ and 
$\V\in\mathbb{R}^{n\times k}$ such that $\M=\U \V^T$ \cite{netflix,kim,candes1,recht,praneeth,keshavan}. Formally, the low rank matrix completion problem for $\M$ with noisy entries over a partial set $\Omega\in[m]\times [n]$ is stated as 
\begin{align}
&(\hat{\U}, \hat{\V}) ~ = ~\underset{\U\in\mathbb{R}^{m\times k} \atop \V\in\mathbb{R}^{n\times k}}{\arg\min}~
\sum\limits_{(i,j)\in\Omega}~(\R_{ij}-\U^{(i)}\V_j)^2,
& \label{LRMC}
\end{align}
where $\R$ is the partially observed noisy version of $\M$. 
The task of inferring missing entries of $\M$ by the above factorization is generally ill-posed unless additional assumptions are made about the structure of $\M$ \cite{candes1}, e.g., 
$\M$ satisfies the incoherence property (see definition (\ref{defn_incoh})) and the entries of 
$\Omega$ are sampled uniformly at random. 
\begin{defn}\cite{candes1}
A rank-$k$ matrix $\M\in\mathbb{R}^{m \times n}$ with SVD given by $\M= \U\Sigma\V^T$ is said to be incoherent with parameter $\mu$ if 
\vskip -20pt
\begin{align*}
&\|\mathbb{P}_{\U}(\e_i)\|_2 \leq \frac{\mu\sqrt{k}}{\sqrt{m}}~\forall~i\in [m], ~~\text{and}~~&\\
&\|\mathbb{P}_{\V}(\e_j)\|_2 \leq \frac{\mu\sqrt{k}}{\sqrt{n}}~\forall~j\in [n].& 
\end{align*}
\label{defn_incoh}
\end{defn}
\vskip -10pt
The optimization in (\ref{LRMC}) is NP-hard \cite{meka2008rank}; a commonly used heuristic 
for approximately solving (\ref{LRMC}) is the alternating minimization approach that keeps 
one of $\U$ and $\V$ fixed and optimizes over the other factor, and then switches and repeats 
the process \cite{praneeth,suriya,hardt2014understanding}. Each of the alternating optimization 
steps is convex and can be solved efficiently. One of the few works that provide a theoretical 
understanding of the convergence properties of alternating minimization based matrix 
completion methods is \cite{praneeth}, where it was shown that for a sufficiently large 
sampling probability of $\Omega$, the reconstruction error can be minimized to an arbitrary 
accuracy. The original noiseless analysis was later extended to a noisy case in \cite{suriya}. 

In a host of applications, factors $\U$, $\V$ or both may 
exhibit structural properties such as sparsity, non-negativity or discreteness. Such applications 
include blind source separation \cite{veen}, gene 
network inference \cite{liao}, and clustering with overlapping clusters 
\cite{arindam}, to name a few. In this work, we consider the rank-one decomposition
of a binary matrix $\M \in \{1,-1\}^{m \times n}$ from its partial observations that are perturbed by 
bit-flipping noise. This formulation belongs to a broader category of non-negative matrix factorization 
\cite{kim} or, more specifically, binary matrix factorization \cite{bmf1,bmf2,dyadic,binary2}. Related 
prior works include \cite{bmf1,bmf2}, which consider decomposition of a binary matrix $\M$ in terms of 
non-binary $\U$ and $\V$, while \cite{dyadic} explores a Bayesian approach to factorizing matrices 
having binary components. The approach in \cite{binary2} constrains $\M$, $\U$ and $\V$ to all
be binary; however, it requires a fully observed input matrix $\M$. On the other hand, 
\cite{slawski2013matrix} considers a factorization of a non-binary $\M$ into a binary and a non-binary factor, with the latter having ``soft'' clustering constraints imposed. As opposed to these works, we aim for approximate factorization in the scenario where all of $\M$, $\U$ and $\V$ are binary, having only limited and noisy access to the entries of $\M$. 

Next, we define the notion of distance between two vectors, which will be used throughout the rest of this paper. 

\begin{defn}\cite{golub}
Given two vectors $\tilde{\u} \in \mathbb{R}^m$ and $\tilde{\w}\in\mathbb{R}^m$, the principal angle  distance between $\tilde{\u}$ and $\tilde{\w}$ is defined as 
\begin{eqnarray*}
\text{dist}(\tilde{\u},\tilde{\w}) = \|\mathbb{P}_{{\u}^{\perp}}(\w)\|_2= \|(\mathbb{I}-\u\u^T)\w\|_2 
= \sqrt{1-(\langle \u,\w\rangle)^2}, 
\end{eqnarray*}
where $\u$ and $\w$ are normalized\footnote{Normalized version of any vector $\x$ is given by $\x/\|\x\|_2$.} forms of $\tilde{\u}$ and $\tilde{\w}$. 
\label{defn_dist}
\end{defn}
\subsection{System model}\label{sec:sys_model}
Let the number of reads carrying information about the haplotypes (after discarding reads which cover no more than one SNP) be $n$. If $m$ denotes the haplotype length ($m \leq n$), then the reads can be organized into an $m \times n$ SNP fragment matrix $\R$, whose $i^{th}$ column $\R_i$ contains information carried by the $i^{th}$ read and whose $j^{th}$ row $\R^{(j)}$  contains information about the $j^{th}$ SNP position on the chromosomes in a pair. Since diploid organisms typically have bi-allelic chromosomes (i.e., only $2$ possible nucleotides at each SNP position), binary labels $+1$ or $-1$ can be ascribed to the informative entries of $\R$, where the mapping between nucleotides and the binary labels follows arbitrary convention.
Let $\Omega$ be the set of entries of $\R$ that carry information about the SNPs; note that the
number of informative entries in each column of $\R$ in much smaller than $m$, reflecting 
the fact
that the reads are much shorter than chromosomes. Let us define the sample probability as $p=\frac{|\Omega|}{mn}$. Furthermore, let us define the operator $P_{\Omega} : \mathbb{R}^{m \times n} \rightarrow \mathbb{R}^{m \times n}$ as
\begin{align}\label{def:Omega}
[P_{\Omega}(\R)]_{ij} & = \begin{cases}
R_{ij},  & (i,j) \in \Omega \\
0, & \text{otherwise}.
\end{cases}
\end{align}

$R_{ij}$ represents the information about the $i^{th}$ SNP site provided by the $j^{th}$ read.
Adopting the convention that $0$'s in column $j$ correspond to SNP positions not covered
by the $j^{th}$ read, $P_{\Omega}(\R)$ becomes a matrix with entries in $\{-1,0,1\}$.  Let $
\mathcal{H} = \{h_1,h_{-1}\}$ be the set of haplotype sequences of the diploid organism under 
consideration, with $h_k \in \{-1,1\}^m, k=1,-1$. Note that the binary encoding of SNPs along 
the haplotypes implies that the haplotypes are binary complements of each other, i.e., 
$h_1 = -h_{-1}$. 

$P_{\Omega}(\R)$ can be thought of as obtained by sampling, with errors, a \textit{rank one} matrix $\M$ with entries from $\{1,-1\}$, given by
\begin{align}
&\M = \hat{\u}^\s (\hat{\v}^\s)^\dagger =  \sigma^\s \u^\s (\v^\s)^\dagger &
\label{eqn:M_svd}
\end{align}
where  $\hat{\u}^\s$ and  $\hat{\v}^\s$ are vectors of lengths $m$ and $n$ respectively, with entries from $\{1,-1\}$, $\u^\s$ and $\v^\s$ are normalized versions of $\hat{\u}^\s$ and $\hat{\v}^\s$, and $\sigma^\s >0$
is the singular value of $\M$. $\hat{\u}^\s$ represents the haplotype $h_1$ or $h_{-1}$ (the choice can be arbitrary) and $\hat{v}_j^\s$ denotes
the membership of $j^{th}$ read, i.e., $\hat{v}_j^\s=k$ if and only if the $j^{th}$ read is sampled from $h_k, k=1,-1$. If $\N$ denotes the sequencing error noise matrix, then the erroneous SNP fragment matrix is given by 
\begin{align}
\R &= \M+\N,~~\text{or}& \nonumber \\
P_{\Omega}(\R) &= P_{\Omega}(\M) + P_{\Omega}(\N).&
\label{eq:haplo_ip_op}
\end{align}
The objective of SIH is to infer $\hat{\u}^\s$ (and $\hat{\v}^\s$) from the data matrix 
$P_{\Omega}(\R)$ which is both sparse as well as noisy. 
\subsection{Noise model}\label{sec:noise_model}
 Let $p_e$ denote the sequencing
error probability. The noise matrix $\N$ capturing the sequencing errors can be modeled as an $m \times n$ matrix with entries in $\{-N_{\max},0,N_{\max}\}$,\footnote{For the labeling scheme adopted in this paper, $N_{\max}=2$.} where each entry is given by 
\begin{align}\label{def:noise}
N_{ij} &=  \begin{cases}
0, ~~\text{w. p.} ~(1-p_e) \\
-2M_{ij}, ~~\text{w. p.}~ p_e. \\
\end{cases}
\end{align}
$\N$ has full rank since the errors occur independently across the reads and SNPs. The 
SVD of $\N$ is given by $\N = \U_N\Sigma_N(\V_N)^\dagger$, where 
$\U_N \in \mathbb{R}^{m \times m}, \V_N \in \mathbb{R}^{n \times m}, 
\Sigma_N = \text{diag}(\sigma_1^N, \sigma_2^N, \ldots, \sigma_m^N)$.  

An important observation about the noise model defined in (\ref{def:noise})
is that it fits naturally into the worst case noise model considered in \cite{suriya,keshavan}. Under this model, the entries of $\N$ are assumed to be distributed arbitrarily,  with the only restriction that there exists an entry-wise uniform upper bound on the absolute value i.e., $|N_{ij}|\leq C$, where $C$ is a constant. This is trivially true for the above formulation of SIH, where $C=N_{\max}$, leading to $\|N\|_F\leq \sqrt{mn}N_{\max}$. With the entries of $\N$ modeled as Bernoulli variables with probability $p_e$,  the following lemma provides a bound on the spectral norm of the partially observed noise matrix $P_{\Omega}(\N)$ and is proved in the 
Appendix~\ref{app:haplo}.  
\begin{lem}\label{lemma:noise}
Let $\N$ be an $m \times n$ sequencing error matrix as defined in (\ref{def:noise}). Let $\Omega$ be the sample set of observed entries and let $p$ be the observation probability. If $p_e$ denotes the sequencing error probability, then with high probability we have 
\vskip 2pt
\begin{align*}
\frac{\|P_{\Omega}(\N)\|_2}{p} & \leq 2N_{\max}p_e m\sqrt{n}. 
\end{align*}
\end{lem}
\section{Single Individual Haplotyping via Alternating Minimization}\label{sec:algo}
As seen in Section~\ref{sec:sys_model}, SIH 
can be formulated as the problem of low-rank factorization of the underlying SNP-fragment 
matrix $\M$,
\begin{eqnarray}
\M = \u {\v}^T.
\label{eq:M_factor}
\end{eqnarray}
To perform the factorization, we optimize the loss function given by 
\begin{align}
&f(\u,\v) = \|P_{\Omega}(\R-\u\v^T)\|_0
=\sum\limits_{(i,j) \in \Omega}~ {\bf 1}_{R_{ij} \neq u_iv_j}, &
\label{eq:haplo_lo_obj}
\end{align}
which is identical to the MEC score associated with the factorization in (\ref{eq:M_factor}). 
However, $\ell_0$-norm optimization problems are non-convex and computationally hard; 
instead, we use a relaxed $\ell_2$-norm loss function 
\begin{align*}
&f(\u,\v) = \|P_{\Omega}(\R-\u\v^T)\|_F^2
=\sum\limits_{(i,j) \in \Omega}~ ({R_{ij} - u_iv_j})^2. &
\end{align*}
Then, the optimization problem can be rewritten as finding $\hat{\u}$ and $\hat{\v}$ such that 
\begin{align*}
&(\hat{\u}, \hat{\v}) =  \underset{{\u} \in \{1, -1\}^m \atop 
{\v} \in \{1, -1\}^n}{\arg \min} f(\u,\v)   
=\underset{{\u} \in \{1, -1\}^m \atop {\v} \in \{1, -1\}^n}{\arg \min}\sum\limits_{(i,j) \in \Omega} ({R_{ij} - u_iv_j})^2.& 
\end{align*}
The above optimization problem can be further reduced 
to a continuous and simpler version by relaxing the binary constraints on $\u$ and $\v$, 
\begin{align}
(\hat{\u}, \hat{\v}) &= \underset{{\u} \in \mathbb{R}^m, 
{\v} \in \mathbb{R}^n}{\arg \min}~~\sum\limits_{(i,j) \in \Omega}~ ({R_{ij} - u_iv_j})^2.& 
\label{eqn:min_relax_prob}
\end{align}
\vskip -5pt

\subsection{Basic alternating minimization for SIH}\label{sec:basic_altmin}
The minimization (\ref{eqn:min_relax_prob}) is a non-convex problem and often eludes globally 
optimal solutions. However, (\ref{eqn:min_relax_prob}) can be solved in a computationally efficient 
manner by using heuristics such as alternating minimization, which essentially alternates
between least-squares solution to $\u$ or $\v$. In other words, the minimization problem boils 
down to an ordinary least-squares update at each step of an iterative procedure, summarized as
\begin{align}
&\hat{\v} ~\leftarrow ~\arg \underset{\v \in \mathbb{R}^n}{\min} ~\sum\limits_{(i,j) \in \Omega}(R_{ij} - {\hat{u}}_i {v}_j)^2, ~~ \text{and}& \label{eq:v_update_basic}\\
&\hat{\u} ~ \leftarrow~ \arg \underset{\u \in \mathbb{R}^m}{\min} ~\sum\limits_{(i,j) \in \Omega}(R_{ij} - {u}_i \hat{v}_j)^2. & \label{eq:u_update_basic}
\end{align}
Once a termination condition is met for the above iterative steps, entries of $\hat{\u}$ are rounded off to $\pm 1$ to give the estimated haplotype vector. 
Following \cite{praneeth}, the basic alternating minimization algorithm for the SIH problem is 
formalized as Algorithm~\ref{algo:basic_altmin}. 
\begin{algorithm}
\caption{SIH via alternating minimization}
\label{algo:basic_altmin}
\begin{algorithmic}
\REQUIRE SNP-fragment matrix ${\R}$, observed set $\Omega$, estimated probability $\hat{p}$ 
\STATE {\bf Power Iteration:} Use power iteration to generate the top singular vector of $P_{\Omega}({\R})/\hat{p}$; denote it by ${\u}^{(0)}$ 
\STATE {\bf Clipping:} Set entries of ${\u}^{(0)}$ greater than $\frac{2}{\sqrt{m}}$ to zero, and then normalize to get ${\hat{\u}}^{(0)}$
\FOR {$t=0,1,2,\ldots, T-1$}
\STATE $\hat{\v}^{(t+1)} \leftarrow \arg \underset{\v \in \mathbb{R}^n}{\min} ~\sum\limits_{(i,j) \in \Omega}(R_{ij} - {\hat{u}}_i^{(t)} {v}_j)^2 $
\STATE $\hat{\u}^{(t+1)} \leftarrow \arg \underset{\u \in \mathbb{R}^m}{\min} ~\sum\limits_{(i,j) \in \Omega}(R_{ij} - {u}_i^{(t)} \hat{v}_j^{(t+1)})^2 $
\ENDFOR
\STATE {\bf Output:} Round-off entries of $\hat{\u}^{(T)}$ to $\pm 1$ to get estimate $\hat{\u}$ of the haplotype vector  
\end{algorithmic}
\end{algorithm}

\noindent\textit{\bf Remark 1:}
Performance of Algorithm~\ref{algo:basic_altmin} depends on the choice of the initial vector $\hat{\u}^{(0)}$. The singular vector corresponding to the topmost singular value of the noisy and partially observed matrix $P_{\Omega}(\R)$ serves as a reasonable starting point since, as shown later in Section~\ref{sec:main_results}, this vector has a small distance\footnote{Please refer to Section~\ref{sec:background_matrix_compl} for a definition of distance measure.} to $\u^{\s}$. However, performing singular value decomposition requires $\mathcal{O}(mn^2)$ operations; therefore, it is computationally prohibitive for large-scale problems typically associated with haplotyping tasks. In practice, the power method is employed to find the topmost singular vector of the appropriately scaled matrix
$P_{\Omega}(\R)$ by iteratively computing vectors $\x^{(j)}$ and $\y^{(j)}$ as
\begin{align}
&\x^{(j)} = P_{\Omega}(\R)\y^{(j-1)}, ~ \y^{(j)} = [P_{\Omega}(\R)]^T\x^{(j)}, ~\forall j=0,1,\ldots & \label{eqn:power_iter}
\end{align}
with the initial $\y^{(0)}$ chosen to be a random vector. Let us assume that the singular values of $P_{\Omega}(\R)$ are $\sigma_1^\prime \geq \sigma_2^\prime \geq \ldots 0$. The power method is guaranteed to converge to the singular vector (say, $\hat{\u}^{(0)}$) corresponding to $\sigma_1^\prime$, provided $\sigma_1^\prime > \sigma_2^\prime$ holds strictly. The convergence is geometric with a ratio $(\sigma_2^\prime/\sigma_1^\prime)^2$. Through successive iterations, the iterate $\x^{(j)}$ gets closer to the true singular vector; specifically,
\begin{align}
\frac{\text{dist}(\x^{(j)},\hat{\u}^0 )}{\|\mathbb{P}_{\hat{\u}^0} (\x^{(j)})\|_2}  & \leq \left( \frac{\sigma_2^\prime}{\sigma_1^\prime}\right)^{2j}  \frac{\text{dist}(\x^{(0)}, \hat{\u}^0 )}{\|\mathbb{P}_{\hat{\u}^0} ( \x^{(0)})\|_2},
\end{align} 
with per iteration complexity of $\mathcal{O}(mn)$ \cite{cai_vikalo} (see Definition \ref{defn_dist} for $\text{dist}(\cdot,\cdot)$). The following lemma provides the complexity of iterations required for the convergence of the described method. 
\begin{lem}[\cite{venkatg}]\label{lemma:power_iter}
Let $\R$ be an $m \times n$ matrix and $\x^{(0)}$ be a random vector in $\mathbb{R}^m$. Let for $\epsilon>0$,  $\bar{U} = \text{span}\left\{\u_i, \forall~ i=2,\ldots \text{~s.t.~} \sigma_i^\prime > (1-\epsilon) \sigma_1^\prime  \right\}$, where $\u_i$'s and $\sigma_i$'s are respectively the left singular vectors and singular values of $\R (in decreasing order of magnitude)$. Then, after $k=\Omega\left(\frac{\log (n/\epsilon)}{\epsilon}\right)$ iterations of the power method, the iterate $\x^{(k)} = \frac{(\R\R^T)^k \x^{(0)}}{\|(\R\R^T)^k \x^{(0)}\|_2}$ satisfies
$ \text{Pr} \left( \mathbb{P}_{\bar{U}_{\perp}}(\x^{(k)}) \geq \epsilon \right) \leq \frac{1}{10}$. 
\end{lem}
\noindent\textit{\bf Remarks 2:} 
It has been shown in \cite{praneeth} that the convergence guarantees for Algorithm~\ref{algo:basic_altmin} can be established, provided the incoherence of the iterates $\hat{\u}^{(t)}$ and $\hat{\v}^{(t)}$ is maintained for iterations $t\geq 0$ (see Definition~\ref{defn_incoh}). To ensure incoherence at the initial step, one needs to threshold or ``clip'' the absolute values of the entries of $\hat{\u}^{(0)}$, as described in Algorithm~\ref{algo:basic_altmin}. Although the singular vector obtained by power iterations minimizes the distance from the true singular vector, it is the clipping step that makes sure that the information contained in $\hat{\u}^{(0)}$ is spread across every dimension instead of being concentrated in only few, much like the true vector $\u^{\star}$ (see Lemma \ref{lemma:init2}). 


\subsection{Binary-constrained alternating minimization}
\label{sec:constrained_altmin}
The updates at each iteration of Algorithm~\ref{algo:basic_altmin} ignore the fact that the underlying 
true factors, namely $\u$ and $\v$, have discrete $\{1,-1\}$ entries; 
instead, the procedure imposes binary constraints on $\u$ and $\v$ at the final step only. This may adversely impact the convergence of alternating minimization; to see this, note that when $\hat{\v}$ is updated in Algorithm~\ref{algo:basic_altmin} according to (\ref{eq:v_update_basic}), 
its $j^{th}$ entry is found as 
\begin{align}
&\hat{v}_j^{(t+1)}  =  \arg \underset{v \in \mathbb{R}}{\min} ~\sum\limits_{i| (i,j) \in \Omega}(R_{ij} - {\hat{u}}_i^{(t)} {v})^2 
=\frac{\sum\limits_{i|(i,j) \in \Omega} R_{ij}\hat{u}_i^{(t)}}{\sum\limits_{i|(i,j)\in \Omega }(\hat{u}_i^{(t)})^2}. &
\label{eq:haplo_iter1}
\end{align}
Clearly, if the absolute value of $\hat{u}_j^{(t)}$ is very large (or very small) compared to $1$ at a given iteration $t$, then, given that $|R_{ij}|=1$ for $(i,j) \in \Omega$, we see from (\ref{eq:haplo_iter1}) that 
the absolute value of $\hat{v}_j^{(t+1)}$ at iteration $t+1$ becomes close to $0$ (or much bigger than 
$1$).
We empirically observe that as the iterations progress, the value of $\hat{v}_j^{(t+1)}$ becomes increasingly bounded away from $\pm 1$, which leads to potential incoherence of the iterates in subsequent iterations. To maintain incoherence,
it is desirable that the entries of $\hat{\u}^{(t)}$ and $\hat{\v}^{(t)}$ remain close to $\pm 1$. 
It is therefore of interest to explore if we can do better by restricting the update steps in the discrete domain; in other words, enforce the discreteness condition in each step, rather than using it at the final step only.  

%
One way of enforcing discreteness is to project the solution of each update onto the set $\{1,-1\}$, 
i.e., impose the inherent binary structure of $\hat{\u}$ and $\hat{\v}$ in (\ref{eq:u_update_basic}) and 
(\ref{eq:v_update_basic}). This leads us to the updates 
\begin{align}
\hat{\v} &\leftarrow \arg \underset{{\v} \in \{1,-1\}^n}{\min} ~\sum\limits_{(i,j) \in \Omega}(R_{ij} - {\hat{u}}_i {v}_j)^2, ~~~\text{and}& \label{eq:v_update_mod}\\
\hat{\u} & \leftarrow \arg \underset{{\u} \in \{1,-1\}^m}{\min} ~\sum\limits_{(i,j) \in \Omega}(R_{ij} - {u}_i \hat{v}_j)^2. & \label{eq:u_update_mod}
\end{align}
Replacing $\u$ and $\v$ updates in Algorithm~\ref{algo:basic_altmin} by (\ref{eq:u_update_mod}) 
and (\ref{eq:v_update_mod}) leads to a discretized version of the alternating minimization algorithm 
for single individual haplotyping, given as Algorithm~\ref{algo:mod_altmin}. Clearly, rounding-off of 
the final iterate is no longer required since the individual iterates are constrained to be binary at 
each step of the algorithm. 

\begin{algorithm}
\caption{SIH via discrete alternating minimization}
\label{algo:mod_altmin}
\begin{algorithmic}
\REQUIRE SNP-fragment matrix ${\R}$, observed set $\Omega$, estimated sequencing
error probability $\hat{p}$.
\STATE {\bf Power Iteration:} Use power iteration to generate the top singular vector of $P_{\Omega}({\R})/\hat{p}$ and denote it by ${\u}^{(0)}$ 
\STATE {\bf Clipping:} Set entries of ${\u}^{(0)}$ greater than $\frac{2}{\sqrt{m}}$ to zero, and then normalize to get ${\hat{\u}}^{(0)}$.
\FOR {$t=0,1,2,\ldots, T-1$}  
\STATE $\hat{\v}^{(t+1)} \leftarrow \arg \underset{\v \in \{1,-1\}^n}{\min} ~\sum\limits_{(i,j) \in \Omega}^{} (R_{ij} - {\hat{u}}_i^{(t)} {v}_j)^2 $ 
\STATE $\hat{\u}^{(t+1)} \leftarrow \arg \underset{\u \in \{1,-1\}^m}{\min} ~\sum\limits_{(i,j) \in \Omega}(R_{ij} - {u}_i^t \hat{v}_j^{(t+1)})^2 $
\ENDFOR 
\STATE {\bf Output:} $\hat{\u}^{(T)}$ is the estimate $\hat{\u}$ of the haplotype vector  
\end{algorithmic}
\end{algorithm}
A closer look at the iterative update of $\hat{\v}$ in Algorithm~\ref{algo:mod_altmin} reveals that the 
update can be written as 
\begin{equation}\label{eqn:step_fn}
{\hat{v}_j}^{(t+1)} = \begin{cases}
1 & \sum\limits_{i|(i,j)\in \Omega}R_{ij}\hat{u}_i^{(t)} \geq 0\\
-1 & \text{otherwise}, ~~~ \forall~j \in [n].
\end{cases}
\end{equation}
Similar update can be stated for $\hat{\u}$. 
The non-differentiability of the update (\ref{eqn:step_fn}), however, makes the analysis of convergence of Algorithm~\ref{algo:mod_altmin} intractable. In order to remedy this problem, the ``hard'' update  in (\ref{eqn:step_fn}) is approximated by a ``soft'' update using a logistic function $f(x) = (e^{x}-1)/(e^{x}+1)$, thus replacing the $\hat{\v}$ and $\hat{\u}$ updates at iteration $t$ in Algorithm~\ref{algo:mod_altmin} by
\begin{align}
\hat{v}_j^{(t+1)} &= f\left(\frac{1}{m}\sum_{i|(i,j)\in \Omega}R_{ij}u_i^{(t)}\right), ~~~ \forall~j~\in~[n], \label{eqn:sigmoid1}\\
\text{and} \nonumber\\
\hat{u}_i^{(t+1)} &= f\left(\frac{1}{n}\sum_{j|(i,j)\in \Omega}R_{ij}v_j^{(t+1)}\right)
, ~~~ \forall~i ~\in ~[m], \label{eqn:sigmoid2}
\end{align}
where $\u^t$ and $\v^t$ are vectors representing normalized $\hat{\u}^t$ and $\hat{\v}^t$. Note that the update steps (\ref{eqn:sigmoid1}) and (\ref{eqn:sigmoid2}) can be represented in terms of the normalized vectors since it  holds that $\text{sign}$ $\left(\sum_{i|(i,j) \in \Omega}~ R_{ij}u_i^t\right)$ = $\text{sign}$ $\left( \sum_{i|(i,j) \in \Omega}~ R_{ij}\hat{u}_i^t\right)$.
The updates (\ref{eqn:sigmoid1}) and (\ref{eqn:sigmoid2}) relax the integer constraints on $\hat{\u}$ and $\hat{\v}$ while ensuring that the values remain in the interval $[1,-1]$.  It should be mentioned here that in the multiple sets of experiments that we have run with synthetic and biological datasets, this approximation did not lead to any noticeable loss of  performance when compared to Algorithm \ref{algo:mod_altmin}; on the other hand, the same allows us to derive an upper bound on the MEC score through an analysis of convergence of the algorithm (see Section~\ref{sec:main_results}).  Algorithm~\ref{algo:sigm_altmin} presents the variant of alternating minimization that relies on soft 
update steps as given by  (\ref{eqn:sigmoid1}) and (\ref{eqn:sigmoid2}).   
\begin{algorithm}
\caption{SIH via discrete alternating minimization with soft updates}
\label{algo:sigm_altmin}
\begin{algorithmic}
\REQUIRE SNP-fragment matrix ${\R}$, observed set $\Omega$, estimated sequencing error probability 
$\hat{p}$.
\STATE {\bf Power Iterations:} Use power iterations to generate the top singular vector of 
$P_{\Omega}({\R})/\hat{p}$ and denote it by ${\u}^{(0)}$.
\STATE {\bf Clipping:} Set entries of ${\u}^{(0)}$ greater than $\frac{2}{\sqrt{m}}$ to zero, and then normalize to get ${\hat{\u}}^{(0)}$.
\FOR {$t=0,1,2,\ldots, T-1$}  
\STATE $\hat{v}_j^{(t+1)} \leftarrow \frac{\exp\left(\frac{1}{m}\sum_{i|(i,j) \in \Omega}R_{ij}u_i^{(t)}\right)-1}
{\exp(\frac{1}{m}\sum_{i|(i,j) \in \Omega}R_{ij}u_i^{(t)})+1}, ~~~\forall~j=1,\ldots,n,$  
\hspace{20mm}
\STATE $\v^{(t+1)}  \leftarrow \hat{\v}^{(t+1)}/\|\hat{\v}^{(t+1)}\|_2$ 
\hspace{20mm}
\STATE $\hat{u}_i^{(t+1)} \leftarrow \frac{\exp\left(\frac{1}{n}\sum_{j|(i,j) \in \Omega}R_{ij}v_j^{(t+1)}\right)-1}
{\exp\left(\frac{1}{n}\sum_{j|(i,j) \in \Omega}R_{ij}v_j^{(t+1)}\right)+1}, ~~~\forall~i=1,\ldots,m, $ 
\hspace{20mm}
\STATE $\u^{(t+1)}  \leftarrow \hat{\u}^{(t+1)}/\|\hat{\u}^{(t+1)}\|_2$ 
\ENDFOR 
\STATE {\bf Output:} Round-off entries of $\hat{\u}^{(T)}$ to $\pm 1$ to get estimate $\hat{\u}$ of the haplotype vector.
\end{algorithmic}
\end{algorithm}


\noindent\textit{\bf Remark 4:}
It is worthwhile pointing out the main differences between the approach considered here and the method 
in \cite{cai_vikalo}. In the latter, the authors propose an alternating minimization based haplotype assembly method by imposing structural constraints on only one of the two factors, namely, the read membership factor. However, for the diploid case considered in this work, use of binary labels allow us to impose similar constraints on both $\u$ and $\v$, thereby leading to computationally efficient yet accurate (as demonstrated in the results section) method outlined in Algorithm~\ref{algo:mod_altmin}. Moreover, the alternating minimization algorithm in \cite{cai_vikalo} is not amenable to performance analysis. Our aim in this paper is to recover (up to noise terms) the underlying true factors and analytically explore relation between the recovery error and the number of iterations required.  

\section{Analysis of Performance}\label{sec:main_results}
We begin this section by presenting our main result on the convergence of  Algorithm~\ref{algo:sigm_altmin}. 
The following theorem provides a sufficient condition for the convergence of this algorithm.  
\begin{thm}\label{theo:main_result}
Let $\hat{\u}^\s \in \{1,-1\}^m$ and $\hat{\v}^\s \in \{1,-1\}^n$ denote the haplotype and read membership vectors, respectively, and let $\R = \M+\N$ denote the observed SNP-fragment matrix where 
$\M = \hat{\u}^\s (\hat{\v}^\s)^T = \u^\s\sigma^\s(\v^\s)^T$, $\N$ is the noise matrix with $N_{\max}$ and $p_e$ as defined in (\ref{def:noise}),
$\u^\s$ and $\v^\s$ are normalized versions of $\hat{\u}^\s$ and
$\hat{\v}^\s$ respectively, and $\sigma^\s$ is the singular value of $\M$. Let $\alpha = n/m \geq 1$ and $\epsilon >0$ be the desired accuracy of reconstruction. 
Assume that each entry of $\M$ is observed uniformly randomly with probability 
\begin{align}
p & >  C\frac{\sqrt{\alpha}}{m\delta_2^2}\log n\log \left( \frac{\|\M\|_F}{\epsilon}\right) \left( p_e+\frac{64}{3} \delta_2\right), \label{eqn:p_cond}
\end{align}
where $\delta_2 \in \left[0, \frac{1}{21}(3.93-C^\p N_{\max}p_e)\right]$ and $C,C^{\p} >0$ are global constants.
Then, after $T = \mathcal{O}(\log (\|\M\|_F/\epsilon))$ iterations of 
Algorithm~\ref{algo:mod_altmin}, the estimate $\hat{\M}^{(T)}=\hat{\u}^{(T)}[\hat{\v}^{(T)}]^T$ with high probability satisfies
\begin{align}
\|\M - \hat{\M}^{(T)}\|_F &\leq  \epsilon + 16\frac{p_e\sigma^\s}{3\delta_2}(2+(2+3N_{\max})\delta_2). \label{eqn:main_result}
\end{align}
\end{thm}
\noindent The following corollary follows directly from Theorem~\ref{theo:main_result}. 
\begin{cor} \label{cor:mec}
Define $\tilde{\M}^{(T)} = \text{sign}\left(\hat{\M}^{(T)}\right)$. Under the conditions of Theorem~\ref{theo:main_result}, the normalized minimum error correction score with 
respect to $\R$, defined as $\tilde{\text{MEC}} = \frac{1}{mn}\|P_{\Omega}(\R-\tilde{\M}^{(T)})\|_0$, 
satisfies 
\begin{eqnarray}
\tilde{\text{MEC}}(\tilde{\M}^{(T)})   \leq ~
\frac{\epsilon}{\sqrt{mn}} + \frac{16p_e}{3\delta_2}(2+(2+3N_{\max})\delta_2) 
 + \frac{1}{\sqrt{mn}}\|P_{\Omega}(\N)\|_F. \label{eqn:mec_result}
\end{eqnarray}
\end{cor}

Theorem~\ref{theo:main_result} and Corollary~\ref{cor:mec} imply that if the sample probability $p$ 
satisfies the condition (\ref{eqn:p_cond}) for a given sequencing error probability $p_e$, then 
Algorithm~\ref{algo:mod_altmin} can minimize the MEC score up to certain noise factors in 
$\mathcal{O}(\log (\|\M\|_F/\epsilon))$ iterations. The corresponding sample
complexity, i.e., the number of entries of $\R$ needed for the recovery of $\M$ is 
$|\Omega|=\mathcal{O}\left(\frac{\sqrt{\alpha}}{\delta_2^2}n\log n\log \left( \frac{\|\M\|_F}{\epsilon}\right) 
\left( p_e+\frac{64}{3} \delta_2 \right) \right)$. Note that compared to (\ref{eqn:main_result}), 
expression (\ref{eqn:mec_result}) has an additional
noise term. This is due to the fact that unlike the loss function $\|\M-\hat{\M}^{(T)}\|_F$ in (\ref{eqn:main_result}), the MEC score of $\tilde{\M}^{(T)}$ is calculated with respect to the observed matrix $P_{\Omega}(\R)$. 
Factor $\log (\|\M\|_F/\epsilon)$ in the expression for sample complexity (\ref{eqn:p_cond}) is due to using independent $\Omega$ samples at each of $T = \mathcal{O}(\log \|\M\|_F/\epsilon)$ iterations \cite{praneeth}. 
This circumvents potentially complex dependencies between successive iterates which are typically hard to analyze \cite{praneethcolt}.  We implicitly assume independent samples of $\Omega$ in each iteration of Algorithm~\ref{algo:sigm_altmin} for the sake of analysis, and consider fixed sample set in our experiments. As pointed out in \cite{praneethcolt}, 
practitioners typically utilize alternating minimization to process the entire sample set 
$\Omega$ at each iteration, rather than the samples thereof. 

The analysis of convergence is based on the assumption that the samples of $\M$ are observed uniformly at random. 
This implies that each read contains SNPs located independently and uniformly at random along the length of the haplotype. 
 In practice, however, the reads have uniformly random starting locations but the positions of SNPs sampled by 
 each read are correlated. 
To facilitate our analysis, the first of its kind for single individual haplotyping, we approximate the practical setting 
by assuming random SNP positions. 


An interesting observation in this context is that sequencing coverage, defined as the number of reads that 
cover a given base in the genome, can conveniently be represented as the product of the sample probability 
$p$ and the number of reads. Then, (\ref{eqn:p_cond}) implies that the required sequencing coverage for 
convergence is $\mathcal{O}\left(\frac{\alpha\sqrt{\alpha}}{\delta_2^2}\log n\log \left( \frac{\|\M\|_F}{\epsilon}\right) 
\left( p_e+\frac{64}{3} \delta_2 \right) \right)$, which is roughly logarithmic in $n$.  


\begin{figure}
\hspace{-0.5cm}
\includegraphics[scale=0.5]{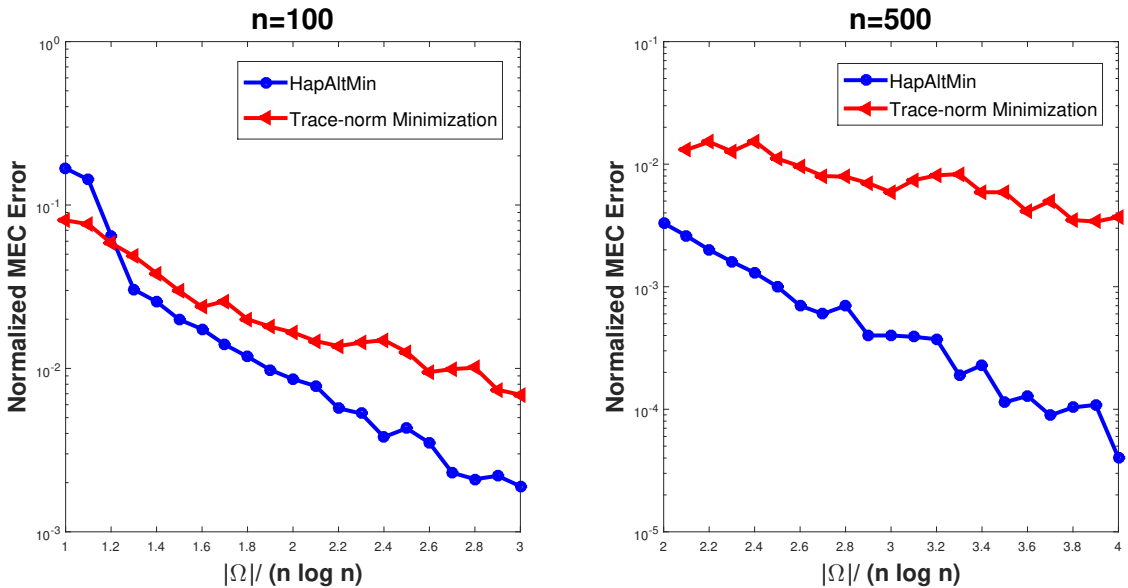}
\caption{A comparison of the normalized minimum error correction score of Algorithm \ref{algo:sigm_altmin} 
and the trace-norm minimization (SVT) method for matrices of dimensions $n=100$ and $n=500$, plotted 
as a function of sample size. $\alpha$ is $2$ and error probability $p_e$ is set to $5\%$. The values shown 
are averaged over $100$ simulation runs.}
\label{fig:hapaltmin_compare}
\end{figure}
\begin{figure}
\hspace{-0.5cm}
\includegraphics[scale=0.5]{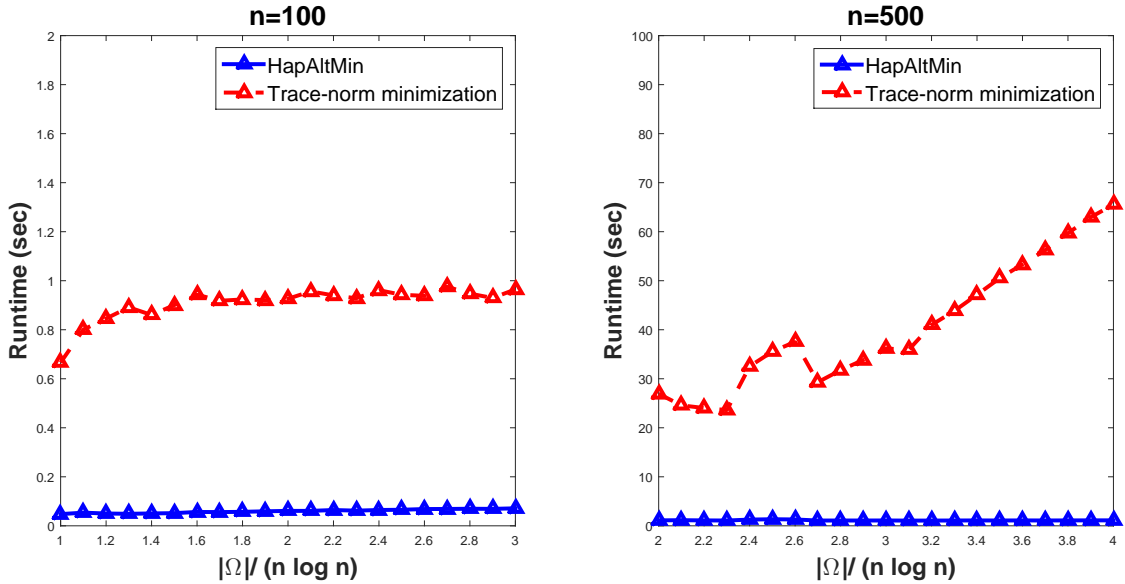}
\caption{A comparison of the runtime of Algorithm \ref{algo:sigm_altmin} and the trace-norm minimization 
(SVT) method for matrices of dimensions $n=100$ and $n=500$, plotted as a function of sample size. 
$\alpha$ is $2$ and error probability $p_e$ is set to $5\%$. The values shown are averaged over $100$ 
simulation runs.}
\label{fig:hapaltmin_runtime_compare}
\end{figure}

In Figure~\ref{fig:hapaltmin_compare} we compare the MEC error rate performance of Algorithm~\ref{algo:sigm_altmin} (denoted as HapAltMin in the figure) with another matrix completion 
approach, namely singular value thresholding (SVT) \cite{svt}; SVT is a widely used trace-norm 
minimization based method. We compare their performance on randomly generated binary rank-one 
matrices of size $50\times 100$ and $250\times 500$, and flip the entries in a uniformly randomly 
chosen sample set $\Omega$ with probability $0.05$.  Both methods are run $100$ times for each 
chosen sample size and error rates are averaged over those runs. Results of the trace-norm minimization 
method are rounded off in the final iterations. Figure~\ref{fig:hapaltmin_compare}  suggests that 
alternating minimization based matrix completion approach performs better than the trace-norm 
minimization based method for both problem dimensions, and the performance gap is wider for 
$n=500$ compared to $n=100$. 
Figure~\ref{fig:hapaltmin_runtime_compare} plots the runtime of the two methods for the same problem 
instances. Trace-norm minimization based method, that was reported in literature as the most accurate version of SVT, is much slower of the two, primarily due to the computationally 
expensive SVD operation at each iteration step.

Next, we present the analysis frameworks for the initial and subsequent iterative steps of Algorithm \ref{algo:sigm_altmin}. 
\subsection{Initialization Analysis}\label{sec:init_analysis}
In order for Algorithm~\ref{algo:sigm_altmin} to converge, a suitable initial starting point close to the ground truth is necessary. In addition to the power iteration step, which gives a singular vector that is close to the true vector $\u^\s$, 
the subsequent clipping step helps retain the incoherence of the same, without sacrificing the closeness property. 
The following lemma uses Theorem \ref{theo:rankkapprox} to establish the incoherence of $\hat{\u}^{(0)}$, and that $\hat{\u}^{(0)}$ remains close to the true left singular vector $\u^{\s}$. 
\begin{lem}\label{lemma:init2}
Let $\u^{(0)}$ be obtained after normalizing the output of the power iteration step in Algorithm~\ref{algo:sigm_altmin}. Let $\u^C$ be the vector obtained after setting entries of $\u^{(0)}$  greater than $2\frac{\mu}{\sqrt{m}}$ to zero. If $\tilde{\u}$ is the normalized $\u^C$, then, with high probability, we have $\text{dist}(\tilde{\u}, \u^{\s}) \leq 1/2$, and 
$ \tilde{\u}$ is incoherent with parameter $4\mu$, where $\mu$ is the incoherence parameter of $\u^{\s}$. 
\end{lem}
\proof
The proof follows directly from Lemma C.2 from \cite{praneeth} and Lemma~2 from  \cite{suriya} after suitably using conditions from Lemma~\ref{lemma:noise} and Theorem~\ref{theo:rankkapprox}. 
\endproof
\subsection{Convergence Analysis}\label{sec:conv_analysis}
Let us denote $\lambda_j = \frac{1}{m}\sum\limits_{i|(i,j) \in \Omega}R_{ij}u_i^{(t)}$.  
The Taylor series expansion of $\hat{v}_j^{(t+1)}$ from (\ref{eqn:sigmoid1}) is given by 
\begin{align}\label{eqn:taylor}
{\hat{v}}_j^{(t+1)} &= \frac{\lambda_j}{2} - \frac{{\lambda_j}^3}{24}+\frac{{\lambda_j}^5}{240}- \ldots, ~~~\forall~j=1,\ldots,n. 
\end{align}
Now, we have 
\begin{eqnarray*}
|\lambda_j| = \left| \frac{1}{m}\sum\limits_{i|(i,j) \in \Omega}R_{ij}u_i^{(t)}\right|  
\leq  \frac{1}{m}\sum\limits_{i|(i,j) \in \Omega}|R_{ij}||u_i^{(t)}|   
\leq  \frac{1}{m}\sum\limits_{i|(i,j) \in \Omega}|R_{ij}|. 
\end{eqnarray*}
Clearly, $\sum\limits_{i|(i,j) \in \Omega}|R_{ij}|\sim \text{Bin}(m,p)$ for a given $j$, implying that the absolute value of $\lambda_j$ is close to the entry-wise observation probability $p < 1$.
This, in turn, implies that in (\ref{eqn:taylor}) all terms with higher powers of $\lambda_j$ are much smaller than the dominant linear term, and the Taylor's series expansion can be written as ${\hat{v}}_j^{(t+1)} \approx  \frac{\lambda_j}{2}  + \epsilon(\lambda_j)$, where the error term $\epsilon(\lambda_j)$ can be bounded as $|\epsilon(\lambda_j)|\leq |\lambda_j|^3/3! \ll 1$ using the Lagrange error bound. Therefore, we approximate the update in (\ref{eqn:sigmoid1}) as 
\begin{equation}\label{eqn:expansion}
\begin{split}
{\hat{v}}_j^{(t+1)} &= \frac{1}{2m}\sum\limits_{i|(i,j) \in \Omega}R_{ij}u_i^{(t)}\\
&=\frac{1}{2m}\sum\limits_{i|(i,j) \in \Omega}(M_{ij}+N_{ij})u_i^{(t)} \\
&=\frac{1}{2m} \sum\limits_{i|(i,j) \in \Omega}\left(\sigma^{\s}u_i^{\s}v_j^{\s} + [\U_N^{(i)}]^T \Sigma_N\V_N^{(j)}\right)u_i^{(t)}  \\
&=\frac{1}{2m} \left(\sigma^{\s}\langle \u^{(t)},\u^{\s}\rangle v_j^{\s} -\left[\sigma^{\s}\langle \u^{(t)},\u^{\s}\rangle v_j^{\s} \right. \right. \\ 
& \hspace{-1mm}\quad \left. \left. - \sigma^{\s}v_j^{\s}\sum\limits_{i|(i,j) \in \Omega}u_i^{(t)}u_i^{\s}\right] + 
 \sum\limits_{i|(i,j) \in \Omega}u_i^{(t)}[\U_N^{(i)}]^T \Sigma_N\V_N^{(j)}\right)  
\end{split}
\end{equation}
for $j=1,\ldots,n$. 

Let us introduce an error vector  $\F \in \mathbb{R}^{n}$ as 
$\F = \sigma^{\s}\B^{-1}\left( \langle  \u^{(t)},\u^{\s} \rangle \B -\C \right)\v^{\s}$, where 
$\B=\frac{1}{p}\mathbb{I}_n$ and $\C\in\mathbb{R}^{n\times n}$ is  diagonal with 
$C_{jj} = \frac{1}{p}\sum\limits_{i|(i,j) \in \Omega}u_i^{(t)}u_i^{\s}, ~\forall j=1,\ldots,n$.  
Furthermore, let us define  a noise vector $N_{\text{res}} \in \mathbb{R}^{n\times 1}$ as
$N_{\text{res}} =\B^{-1}\C^N\S^N\v^N$, where the quantities are as follows:
\begin{itemize}
\item[-]
$\C^N = [\C_1^N~\C_2^N \cdots \C_m^N] \in\mathbb{R}^{n\times nm}$ where $(\C_q^N)_{jj} = \frac{1}{p}\sum\limits_{i|(i,j) \in \Omega} u_i^{(t)}\U_{iq}^N, ~~\forall q\in[m],\forall j\in[n]$;
\item[-] $\S^N \in \mathbb{R}^{nm\times nm}$ is a diagonal matrix given by $\S^N = \text{diag}\left( \sigma_1^N\mathbb{I}_n,\ldots,\sigma_m^N\mathbb{I}_n\right)$;
\item[-] $\v^N = \left[ (\v_1^N)^T~ (\v_2^N)^T~ \ldots ~(\v_m^N)^T\right]^T \in\mathbb{R}^{nm\times 1}$ where $\v_j^N \in \mathbb{R}^{n}$ is the  $j^{th}$ column of $\V_N$. 
\end{itemize}
We also define $C_{jj}^N = \left[ (\C_1^N)_{jj}~(\C_2^N)_{jj}~\ldots~(\C_m^N)_{jj}\right]$ for any given $j\in [n]$. Using the above definitions, (\ref{eqn:expansion}) can be written as 
\begin{eqnarray}
\hat{v}_j^{(t+1)} =  
\frac{1}{2m}\left[ \sigma^{\s}\langle \u^{(t)},\u^{\s} \rangle v_j^{\s} - F_{j} +   \frac{1}{B_{jj}}C_{jj}^N \Sigma_N \V_N^{(j)} \right].    \label{eqn:vj_expan}
\end{eqnarray}

\noindent Therefore, using vector-matrix notation, the update of $\hat{\v}$ can be written as 
\begin{align}
{\hat{\v}}^{(t+1)} & =  \frac{1}{2m} \left[\sigma^{\s}\langle \u^{(t)},\u^{\s}\rangle \v^{\s} - \F + N_{\text{res}}\right] & \label{eqn:v_expan} \\
& =  \frac{1}{2m} \left[\M^T\u^{(t)} - \F + N_{\text{res}}\right]. &  \nonumber
\end{align}
Recalling (\ref{eqn:power_iter}), one can identify that $\M^T\u^{(t)}$ in the above expression is the update term in the power iteration applied to the true matrix $\M$. Therefore, the update described in (\ref{eqn:v_expan}) is essentially power iteration of $\M$ that would have led to the true singular vector $\u^\s$ except that it is perturbed by error terms due to incomplete observations and sequencing noise ($\F$ and $N_{\text{res}}$ in the above expression, respectively). This observation leads to the analysis approach where appropriate upper bounds to the aforementioned errors terms are derived. 
The proof of convergence presented in this work follows the framework adopted in \cite{praneeth,suriya}, and consists of an inductive analysis based on establishing the guarantee that, given $\u^{(t)}$ is incoherent and has a small distance in terms of principal angle with respect to $\u^\s$, the subsequent iterate $\v^{(t+1)}$ is also incoherent with identical parameter, and is closer to $\v^\s$ by at least a constant factor. This statement is formally expressed using   Theorem~\ref{theo:sec_result} and Lemma~\ref{lemma:incoherence} for the distance and incoherence conditions, respectively. 

\begin{thm}\label{theo:sec_result}
Under the assumptions of Theorem~\ref{theo:main_result}, the $(t+1)^{th}$ iterates $\hat{\u}^{(t+1)}$ and $\hat{\v}^{(t+1)}$ of Algorithm~\ref{algo:sigm_altmin} satisfy with high probability 
\begin{align*}
\text{dist}(\hat{\v}^{(t+1)}, \v^{\s})  &\leq \frac{1}{4} \text{dist}(\hat{\u}^{(t)}, \u^{\s}) + \frac{\mu_1 p_e}{\delta_2}, ~~ \text{and} \\
\text{dist}(\hat{\u}^{(t+1)}, \u^{\s})  &\leq \frac{1}{4} \text{dist}(\hat{\v}^{(t+1)}, \v^{\s})+ \frac{\mu_1 p_e}{\delta_2}, ~0\leq t\leq T-1, 
\end{align*}
where $\mu_1$ is the incoherence parameter of the intermediate iterates $\hat{\u}^{(t)}$ and $\hat{\v}^{(t)}$. 
\end{thm}
We defer the proofs of both Theorem~\ref{theo:main_result} and Theorem~\ref{theo:sec_result} to Appendix~\ref{app:haplo}.
Theorem~\ref{theo:sec_result} is a starting point for proving Theorem~\ref{theo:main_result}, and establishes 
a geometric decay of the  distance between the subspaces spanned by $(\hat{\u}^{(t)}$, $\hat{\v}^{(t)})$ and 
$(\u^{\s}$, $\v^{\s})$ respectively. Next, we present a corollary based on the findings of Theorem~\ref{theo:sec_result}. 
\begin{cor}\label{cor:sec_result}
Under the assumptions of Theorem~\ref{theo:sec_result}, at the end of 
$T = \mathcal{O}(\log \frac{\|\M\|_F}{\epsilon})$ iterations, it holds that
\begin{align*}
\text{dist}(\hat{\v}^{(T+1)}, \v^{\s})  &\leq \frac{1}{2}\frac{\epsilon}{\|\M\|_F} + \frac{4\mu_1 p_e}{3\delta_2}, ~~ \text{and} 
\\
\text{dist}(\hat{\u}^{(T+1)}, \u^{\s})  &\leq \frac{1}{2}\frac{\epsilon}{\|\M\|_F} + \frac{4\mu_1 p_e}{3\delta_2}. 
\end{align*}
\end{cor}
\proof
It follows from Theorem~\ref{theo:sec_result} that, after $T$ iterations, 
\begin{equation*}
\begin{split}
\text{dist}(\hat{\u}^{(T)}, \u^{\s}) & \leq \frac{1}{4} \left( \frac{1}{4}\text{dist}(\hat{\u}^{(T-1)}, \u^{\s})  +   \frac{\mu_1 p_e}{\delta_2} \right) +  \frac{\mu_1 p_e}{\delta_2}
\\
& = \frac{1}{16}\text{dist}(\hat{\u}^{(T-1)}, \u^{\s})  + \frac{5\mu_1 p_e}{4\delta_2}
\\
& \vdots
\\
&\leq \frac{1}{16^T}\text{dist}(\hat{\u}^{(0)}, \u^{\s})  + \frac{5\mu_1 p_e}{4\delta_2}\left(1+\frac{1}{16}  
\frac{1}{16^2}+  
 + \cdots \text{(T terms)}\right) 
\\
& \overset{\zeta_1}{\leq} \frac{1}{2}\frac{1}{16^T} + \frac{4\mu_1 p_e}{3\delta_2}
\\
& \leq \frac{1}{2}\frac{\epsilon}{\|\M\|_F} + \frac{4\mu_1 p_e}{3\delta_2}, ~ \text{as}~ 
T=\mathcal{O}\left(\log \frac{\|\M\|_F}{\epsilon}\right)
\end{split}
\end{equation*}
where $\zeta_1$ follows from the fact that $\text{dist}(\hat{\u}^{(0)},\u^\s)\leq 1/2$ (see Lemma~\ref{lemma:init2}).  
\endproof
Corollary~\ref{cor:sec_result} is used to complete the proof of Theorem \ref{theo:main_result} in Appendix \ref{app:haplo}.   The following lemma states the incoherence condition; we defer its proof to Appendix \ref{app:haplo}.
\begin{lem}\label{lemma:incoherence}
Let $\M, \N$, $p$, $\Omega$ be defined as in Theorem~\ref{theo:main_result}. Let $\u^{(t)}$ be the unit vector obtained at the $t^{th}$ iteration of Algorithm~\ref{algo:sigm_altmin} with incoherence parameter $\mu_1=8\mu$. Then, with probability greater than $1-1/n^3$, the next iterate $\v^{(t+1)}$ is also $\mu_1$ incoherent. 
\end{lem}
\noindent\textit{\bf Remark 5:} In our analysis of the matrix factorization approach to single individual 
haplotyping, we adopted techniques proposed by \cite{praneeth,suriya}; note, however, that the scope of analysis 
in the current paper goes beyond the prior works. In particular, the authors of \cite{praneeth} considered a noiseless 
case of matrix factorization and did not impose structural constraints on the iterates. Their work was extended 
to a noisy case in \cite{suriya} for a similar unconstrained setting. Additionally, \cite{suriya} did not exploit statistical 
properties of the noise, except the entry-wise upper bound, whereas the present work uses the bit-flipping model as discussed 
in Section \ref{sec:noise_model}, which allows us to characterize the dependence of the performance on the 
sequencing error probability $p_e$.

\section{Results and Discussions}\label{sec:experiments}
We begin this section by stating the metrics for evaluating performance of our single individual
hapolotyping algorithm. 
\subsection{Performance Metrics}\label{sec:perf_metric}
A widely used metric for characterizing the quality of single individual haplotyping is the minimum 
error correction (MEC) score. This metric captures the smallest number of entries of $P_{\Omega}(\R)$ 
which need to be changed from $1$ to $-1$ and vice versa so that $P_{\Omega}(\R)$ can be 
interpreted as a noiseless version of $P_{\Omega}(\M)$. Essentially, it is the most likely number of 
sequencing errors, defined for diploids as
\begin{align}
\text{MEC} &= \sum\limits_{i=1}^n~\min \left( D(r_i,\hat{h}_1), D(r_i,\hat{h}_{-1})\right),
\label{eq:MEC_defn}
\end{align}
where $D(r_i,\hat{h}_j)$ denotes the generalized Hamming distance between read $r_i$ (regarded as $m$ length vector in $\{1,-1,\text{`-'}\}$) and the estimated parent haplotypes $\hat{h}_k, ~k=1, -1$. This, in turn, is defined as 
$D(r_i,\hat{h}_j) = \sum_{j=1}^m~d(r_{i,j}, \hat{h}_{k,j})$, where 
\begin{align}
d(x,y) = \begin{cases} 1, & ~\text{if}~x\neq \text{`-' ~and ~} y\neq \text{`-' ~and ~} x\neq y \\
0, &~\text{otherwise},
\end{cases}
\label{eq:ham_dist_defn}
\end{align} and 
$r_{i,j}$ and $\hat{h}_{k,j}$ denote the $j^{th}$ entries of $r_i$ and $\hat{h}_k$, respectively.  

The MEC score 
is a relevant and most commonly studied performance metric for single individual haplotyping 
\cite{schwartz2010theory}, and critically important for experimental data where the ground truth is not 
known in advance. It is also a proxy for the most meaningful haplotype assembly metric referred to as
\textit{reconstruction rate}. 
Recall that $\mathcal{H} = \{h_1,h_{-1}\}$ denotes the set of true haplotypes.  
Then the reconstruction rate of $\hat{\mathcal{H}}= \{\hat{h}_1,\hat{h}_{-1}\}$ with respect to $\mathcal{H}$ is defined as \cite{geraci}
\begin{eqnarray}
\mathcal{R}_{\mathcal{H},\hat{\mathcal{H}}} = 1- \frac{\min \{ D(h_1,\hat{h}_1+D(h_{-1},\hat{h}_{-1}), D(h_1,\hat{h}_{-1})+D(h_{-1},\hat{h}_1) \}}{2m},
\end{eqnarray} 
where $D(h_i,h_j)$ denotes the generalized Hamming distance between the haplotype pair $h_i$ and $h_j$. 
%
%


\subsection{Experiments}\label{sec:exp_results}

In this section, for convenience we refer to our algorithm for single individual haplotyping as \textit{HapAltMin} .  
All of the methods described here were run on a Linux OS desktop with 3.07 GHz CPU and 8 GB RAM on an 
Intel Core i7 880 Processor. 

We first tested our algorithm on the experimental dataset containing Fosmid pool-based NGS 
data for HapMap trio child $NA12878$ \cite{duitama}. The Fosmid dataset is characterized 
by very long fragments, high SNP to read ratio, and low sequencing coverage of about 3X, consisting of around $1,342,091$ SNPs spread across $22$ chromosomes. We compare the performance of HapAltMin with 
the structurally-constrained gradient descent (SCGD) algorithm of \cite{cai_vikalo} and one 
more recent SIH software ProbHap \cite{probhap}, which was shown to be superior to several 
prior methods on this dataset \cite{bansal_hapcut,duitama,mixsih}.  Table~\ref{table:exp} shows 
the MEC rate (average number of mismatches per SNP position across the reads) and runtimes 
for all $22$ chromosomes. As seen there, our HapAltMin outperforms other methods for majority 
of the chromosomes shown; it is second best in terms of runtime (behind SCGD).

\begin{table}[h]
\caption{MEC rates and runtimes on HapMap sample NA12878 dataset.}
\label{table:exp}
\centering
\begin{tabular}{|C{0.7cm}|C{1.5cm}|C{1.4cm}|C{1.5cm}|C{1.4cm}|C{1.5cm}|C{1.4cm}|}
\hline\hline 
{\bf Chr} &  \multicolumn{2}{|c|}{{\bf HapAltMin}} & \multicolumn{2}{|c|}{{\bf SCGD}}  & \multicolumn{2}{|c|}{{\bf ProbHap}} \\ \hline
& {\bf MEC}  & {\bf time(s)} & {\bf MEC}  & {\bf time(s)} & {\bf MEC}  & {\bf time(s)} \\ \hline
\hline
1&  0.034 & 65.0 & 0.04 & 44.2& 0.058 & 87.7 \\\hline  
2&  0.035  & 71.6 & 0.035 & 49.5 & 0.055 & 88.9   \\ \hline
3& 0.034 & 61.1 & 0.036 & 41.5 & 0.057 & 84.3 \\ \hline
4& 0.029 & 60.7 & 0.034 & 41.8 & 0.053 & 67.1\\ \hline
5&  0.032  & 52.9 & 0.036 & 39.9 & 0.054 & 64.6 \\ \hline
6 & 0.038 &  34.7 &  0.037 & 27.9&  0.050 & 53.4 \\ \hline
7 & 0.038 &26.4&  0.035 &25.05& 0.055 & 40.8\\ \hline
8 & 0.033 &24.3&  0.034 &23.9& 0.05 & 42.8\\ \hline
9 & 0.036 &21.9&  0.037 &17.6& 0.052& 45.2\\ \hline
10 & 0.036 & 24.7&  0.037 &21.0& 0.053& 44.4 \\ \hline
11 & 0.034 & 24.7&  0.038 &20.8& 0.055& 39.5\\ \hline
12 & 0.037 &23.5&  0.037 & 20.2& 0.057& 38.9\\ \hline
13 & 0.039 &14.6&  0.035 &15.6& 0.053& 26.4\\ \hline
14 & 0.035 &16.6&  0.039 &13.7& 0.055& 27.4\\ \hline
15 & 0.038 & 14.1&  0.041 &11.9& 0.056& 26.5\\ \hline
16 & 0.046 &20.3&  0.0405 & 12.2& 0.051& 36.5\\ \hline
17 & 0.048 &15.3&  0.046 &11.1& 0.061& 27.4\\ \hline
18 & 0.033 &12.2&  0.037 &11.8& 0.053& 24.4\\ \hline
19 & 0.052 &12.8&  0.046 &9.0& 0.063& 19.8\\ \hline
20& 0.044 & 18.1 & 0.044 & 13.0 & 0.055 & 30.9\\ \hline
21&  0.035 & 11.5& 0.041 & 8.5 & 0.051 & 15.6\\ \hline
22&  0.054 & 11.7& 0.055 & 8.6 & 0.061 & 31.4 \\ \hline
\end{tabular}
\vspace{-0.05in}
\end{table}

\begin{table}[!t]
\caption{MEC rate comparison of Algorithm~\ref{algo:mod_altmin} and Algorithm~\ref{algo:sigm_altmin}.}
\label{table:comparison}
\centering
\begin{tabular}{|C{2cm}|C{3.2cm}|C{3.2cm}|}
\hline\hline 
{\bf Chr} &  {\bf Algorithm~\ref{algo:mod_altmin} (``Hard'' updates)} &  {\bf Algorithm~\ref{algo:sigm_altmin} (``Soft'' updates)}\\ \hline
\hline
1    & 0.0369 & 0.0339 \\ \hline
2  & 0.0357 & 0.035\\ \hline
3  & 0.0345& 0.0335 \\ \hline
20  & 0.0447 & 0.0446\\ \hline
21  & 0.0367 & 0.035\\ \hline
22  & 0.0568& 0.0535 \\ \hline
\end{tabular}
\vspace{-0.05in}
\end{table}
We further compare the MEC performance of Algorithm \ref{algo:sigm_altmin} with that of the discrete version (Algorithm~\ref{algo:mod_altmin}) in Table~\ref{table:comparison} for first $3$ and last $3$ of the $22$ chromosomes from the Fosmid dataset. As can be seen from Table~\ref{table:comparison}, the modified algorithm performs better than the discrete alternating minimization algorithm, as indicated in the discussion in Section~\ref{sec:algo}. 

Next, we focus on the evaluation of performance on simulated dataset using reconstruction rate metric. For this purpose, we use
a widely popular standard benchmarking dataset from \cite{geraci} which also provides the true haplotypes used to generate the read data.  The dataset contains reads at a sequencing error rate with values in the set $\{0.0, 0.1, 0.2, 0.3\}$, and a depth of coverage in the set $\{3X, 5X, 8X, 10X\}$.
Reconstruction rate of our algorithm is 
compared with that of SCGD \cite{cai_vikalo} and two more recent SIH methods known as HGHap \cite{hghap} and MixSIH \cite{mixsih}. In particular, \cite{hghap} is chosen for performance comparison since it has been shown to outperform a number of existing SIH methods such as \cite{levy,chen2008linear,bansal_hapcut,genovese2007fast,panconesi2004fast}.
A comparison with ProbHap is not shown for this data since it reconstructed haplotypes with a large fraction of SNPs missing (and therefore has inferior performance compared to the methods
used in the comparisons).
The results, shown in Table~\ref{table:simu}, are obtained by averaging over $100$ simulation runs for each combination of sequencing error rate and sequencing coverage, for a haplotype length of $700$ base pairs and pairwise hamming 
distance $0.7$.  As evident from the results, our method is either the best or the second best in all of the scenarios. HapAltMin performs particularly well in the more realistic error range of $0.0-0.2$ and is marginally inferior to MixSIH only for higher sequencing error values. 
\begin{table}[!t]
\caption{Reconstruction rate comparison on simulated data. Boldface values indicate best performance.}
\label{table:simu}
\centering
\begin{tabular}{|C{1.7cm}|C{1.2cm}|C{2.3cm}|C{2.1cm}|C{2.1cm}|C{2.1cm}|}
\hline\hline 
{\bf Error Rate} & {\bf Cov.}  & {\bf HapAltMin} & {\bf SCGD}  & {\bf HGHap} & {\bf MixSIH} \\ 
\hline\hline 
0.0 & 3X & {\bf 1}  & 0.983 & 0.934 	& 
0.776
\\ \hline
0.0 & 5X & {\bf 1}  & 0.976 &  0.989	& 0.923
\\\hline
0.0 & 8X &  {\bf 1}   &  0.999 & 0.994		& 0.995
\\\hline
0.0 & 10X &  {\bf 1} &  0.999 & 0.999 	& {\bf 1}
\\\hline
0.1 & 3X & {\bf 0.935}  & 0.869 & 0.934 	& 
0.775
\\ \hline
0.1 & 5X & 0.979  & 0.951 & {\bf 0.990}		& 0.942
\\\hline
0.1 & 8X &  {\bf 0.996}   & {\bf 0.996} & 0.987		& 0.972
\\\hline
0.1 & 10X & {\bf 0.999}  & {\bf 0.999} & 0.997 	& 0.993
\\\hline
0.2 & 3X & {\bf 0.735}   & 0.677 & 0.677		& 0.68
\\\hline
0.2 & 5X & 0.864   & 0.785 & {\bf 0.91}		& 0.774
\\\hline
0.2 & 8X & {\bf 0.943}   & 0.899 & 0.884	& 0.932
\\\hline
0.2 & 10X & 0.966  & 0.934 & 0.894	& {\bf 0.969}
\\\hline
0.3 & 3X & 0.555   & 0.527 & 0.592		& {\bf 0.65}
\\\hline
0.3 & 5X & 0.595   & 0.524 &  0.621		& {\bf 0.667}
\\\hline
0.3 & 8X & 0.68   & 0.518 & 0.646	& {\bf 0.714}
\\\hline
0.3 & 10X & 0.723  & 0.58 & 0.696 & {\bf 0.751}
\\\hline
\end{tabular}
\vspace{-0.05in}
\end{table}

\section{Conclusion}
Motivated by the single individual haplotyping problem from
computational biology, we proposed and analyzed a binary-constrained variant of the 
alternating minimization algorithm for solving the rank-one matrix factorization problem. 
We provided theoretical guarantees on the performance of the algorithm and analyzed 
its required sample probability; the latter has important implications on experimental 
specifications, namely, sequencing coverage. Performance of haplotype reconstruction 
is often expressed in terms of the minimum error correction score; we establish theoretical 
guarantees on the achievable MEC score for the proposed binary-constrained alternating 
minimization. Experiments with a HapMap sample NA12878 dataset as well as those with 
a widely used benchmarking simulated dataset demonstrated efficacy of our algorithm.


%

\appendices
\section{Proofs of Lemmas and Theorems}\label{app:haplo}
\subsection{Preliminaries}

The following property and lemma are well-known classical results that will be useful in the
forthcoming proofs.

\begin{property} For a given matrix $\M \in \mathbb{R}^{m \times n}$ of rank $k$, the following relations hold between the 2-norm, the Frobenius norm and the entry-wise $\ell_1$ norm of $\M$:
\begin{itemize}
\item $\|\M\|_2 \leq \|\M\|_F \leq \sqrt{k}\|\M\|_2 $
\item $ \|\M\|_F \leq \|\M\|_1 \leq \sqrt{mn}\|\M\|_F$ 
\end{itemize}
\end{property}
\begin{lem}[Bernstein's inequality]\label{lemma:bernstein}
Let $X_1, X_2, \ldots, X_n$ be independent random variables. Also, let 
$|X_i|\leq L \in \mathbb{R}~\forall  i$ w.p. 1. Then it holds that
\begin{eqnarray*}
Pr \left(\left|\sum\limits_{i=1}^{n}X_i - \sum\limits_{i=1}^{n}\mathbb{E}[X_i]\right|> t\right) \leq  
2\exp\left(-\frac{t^2/2}{Lt/3+\sum\limits_{i=1}^{n}\text{Var}(X_i)}\right).
\end{eqnarray*}
\end{lem}
The following theorem from \cite{keshavan} provides an upper bound on the error between the true matrix $\M$ and the best rank-$k$ approximation of the noisy and partially observed version of $\M$, and is used in the proof of Lemma~\ref{lemma:init2}.  
\begin{thm}{\cite{keshavan}}\label{theo:rankkapprox}
Let $\R=\M+\N$, where $\M$ is an $m \times n$ $\mu$-incoherent matrix with rank $k$ ($m \leq n$) and the indices in the sampling set $\Omega \in [m] \times [n]$ 
are chosen uniformly at random.
Let $\alpha=n/m$, $|M_{ij}| \leq M_{\max}$ and $p$ be the sampling probability. Furthermore, from the SVD of ~$\frac{1}{p}P_{\Omega}(\R)$,
we get a rank-$k$ approximation given by $[P_{\Omega}(\R)]_k =
   {\U}^0{\Sigma}^0 ({\V}^0)^T$. Then there exists numerical constants $C$ and $C^\p$ such that, with probability greater than $1-\frac{1}{n^3}$, we have 
   \begin{eqnarray*}
   \frac{1}{\sqrt{mn}}\|\M-[{P}_{\Omega}(\R)]_k\|_2 \leq CM_{\max}\left(\frac{m\alpha^{3/2}}{|\Omega|} \right)^{1/2}   + \frac{C^\p m\sqrt{\alpha}}{|\Omega|}\|{P}_{\Omega}(\N)\|_2.
   \end{eqnarray*}   
   \end{thm}  
\subsection{Induction Proofs}
\begin{lem}\label{lemma:F_bound}
Let $\M,\N,\Omega$ and $\u^{(t)}$ be defined as  in Algorithm \ref{algo:sigm_altmin}. Then, with high probability we have 
\begin{align*}
\|\F\|_2 \leq \sigma^{\s}\delta_2\sqrt{1-{\langle \u^{(t)},\u^{\s}\rangle} ^2}.
\end{align*}
\end{lem}  
\proof
From the definition of $\F$ stated in Section~\ref{sec:conv_analysis}, we have
\begin{align}\label{eqn:F_definition}
\|\F\|_2 & \leq \sigma^{\s}\|\B^{-1}\|_2\|\left(\C-{\langle \u^{(t)},\u^{\s}\rangle}\B\right)\v^{\s}\|_2.
\end{align}
Since $\B$ is a diagonal matrix, $\|\B^{-1}\|_2 = \frac{1}{\min_i B_{ii}}=p \leq 1$. 
Let $\x\in \mathbb{R}^n$ be such that $\|\x\|_2=1$. Then, for all such $\x$, 
\begin{equation*}
\begin{split}
&\x^T \left({\C-\langle \u^{(t)},\u^{\s}\rangle}\mathbb{I}\right)\v^{\s}  
\\
& \quad = \frac{1}{p}\sum\limits_{j}x_jv_j^{\s}
\left( \sum\limits_{i|ij\in\Omega}u_i^{(t)}u_i^{\s} - \langle \u^{(t)},\u^{\s}\rangle\right)  
\\
& \quad \leq \frac{1}{p}\sum\limits_{ij\in\Omega} x_jv_j^{\s} \left(u_i^{(t)}u_i^{\s}  - {\langle \u^{(t)},\u^{\s}\rangle(u_i^{(t)})^2}\right)
\\
& \quad  \overset{\zeta_1}{\leq} \frac{C}{p}\sqrt{np} \sqrt{\sum\limits_{j}x_j^2(v_j^{\s})^2}\sqrt{\sum\limits_{i} \left(u_i^{(t)}u_i^{\s}  - {\langle \u^{(t)},\u^{\s}\rangle(u_i^{(t)})^2}\right)^2} 
\\
& \quad \overset{\zeta_2}{\leq} \frac{C}{p}\sqrt{np} \sqrt{\frac{\mu_1^2}{n}\sum\limits_{j}x_j^2}~\sqrt{\sum\limits_{i}  (u_i^{(t)})^2  \left( u_i^{\s}  - {\langle \u^{(t)},\u^{\s}\rangle u_i^{(t)}}\right)^2} 
\\
& \quad  \overset{\zeta_3}{\leq} \frac{C}{p}\frac{\sqrt{np}\mu_1^2}{\sqrt{mn}}\sqrt{1-{\langle \u^{(t)},\u^{\s}\rangle} ^2} 
\\
& \quad  \leq \delta_2\sqrt{1-{\langle \u^{(t)},\u^{\s}\rangle} ^2}, \qquad \text{if}~~ p\geq C^{\prime}\frac{\mu_1^4}{m\delta_2^2}, 
\end{split}
\end{equation*}
where $C^{\prime}=C^2>0$ is a global constant and $\zeta_1$ follows from Lemma \ref{lemma:normbound} (which 
imposes the condition $p\geq C\frac{\log n}{m}$) and $\zeta_2$ follows from the incoherence of $\v^{\s}$, and $\zeta_3$ from that of $\u^{(t)}$. Then, 
\begin{align*}
\|\left({\C-\langle \u^{(t)},\u^{\s}\rangle}\B\right)\v^{\s}\|_2 &= \underset{\|\x\|_2=1}{\max}~ \x^T\left({\C-\langle \u^{(t)},\u^{\s}\rangle}\B\right)\v^{\s}& \\
&\leq \delta_2\sqrt{1-{\langle \u^{(t)},\u^{\s}\rangle}^2}.&
\end{align*}
Hence, the lemma follows from (\ref{eqn:F_definition}) if $p\geq C^{\prime}\frac{\mu_1^4\log n}{m\delta_2^2}$. 
\endproof
\begin{lem}{(\cite{praneeth})}\label{lemma:normbound}
Let $\Omega \in [m]\times [n]$ be a set of indices sampled uniformly at random with sampling probability $p$
 that satisfies $p \geq C\frac{\log n}{m}$. Then with probability $\geq 1-\frac{1}{n^3}$ $\forall \x\in\mathbb{R}^m, \forall \y \in \mathbb{R}^n$ such that $\x$ satisfies $\sum_i x_i=0$, it holds that $\sum\limits_{ij\in\Omega}x_iy_j \leq C\sqrt{\sqrt{mn}p}\|\x\|_2\|\y\|_2$, where $C>0$ is a global constant.  
\end{lem}
\begin{lem}\label{lemma:noise_term_bound}
Let $\M,\N,\Omega$ and  $p_e$ be defined as before. Then with high probability it holds that
\begin{align*}
\|N_{\text{res}}\|_2 \leq  2N_{\max}\mu_1 p_e\sqrt{mn}. 
 \end{align*}
\end{lem}
\proof
The proof follows from Lemma B.3, \cite{suriya} for the case $k=1$, and by noting the fact that $\|\B^{-1}\|_2 \leq 1$, and  from the observation that  $\frac{\|\mathcal{P}_{\Omega}(\N)\|_2}{p}  \leq 2N_{\max}p_e\sqrt{mn}, ~
\text{with high probability}$ (see Lemma~\ref{lemma:noise}). 
\endproof
\begin{lem}\label{lemma:vnorm_bound}
Let $\F$, $N_{\text{res}}$ and $\u^{(t)}$ be defined as in (\ref{eqn:v_expan}). Then  we have
\begin{align*}
\|{\hat{\v}}^{(t+1)}\|_2\geq \frac{1}{2m}\left( \sigma^{\s}\sqrt{1-\text{dist}^2(\u^{\s},\u^{(t)})} -\|\F\|_2 - \|N_{\text{res}}\|_2 \right).
\end{align*} 
\end{lem}
\proof
\begin{align*}
 2m\|{\hat{\v}}^{(t+1)}\|_2 & = \|\sigma^{\s}\langle \u^{(t)},\u^{\s}\rangle \v^{\s}-\F+N_{\text{res}}\|_2 &\\
& \overset{\zeta_1}{\geq} \|\sigma^{\s}\langle \u^{(t)},\u^{\s}\rangle \v^{\s}\|_2 - \|\F-N_{\text{res}}\|_2& \\
&\overset{\zeta_2}{\geq} \|\sigma^{\s}\langle \u^{(t)},\u^{\s}\rangle \v^{\s}\|_2 - \|\F\|_2 -\|N_{\text{res}}\|_2 & \\
& = \sigma^{\s}\sqrt{1-\text{dist}^2(\u^{\s},\u^{(t)})^2} -\|\F\|_2 - \|N_{\text{res}}\|_2,  & 
\end{align*}
where both $\zeta_1$ and $\zeta_2$ follow from the reverse triangle inequality for vectors. 
\endproof
\begin{lem}\label{lemma:noise_entry_bound}
Under the conditions of Theorem~\ref{theo:main_result}, with probability greater than $1-1/n^3$ it holds
that for a given $j \in [n]$, 
\begin{align*}
\left|\frac{1}{B_{jj}}C_{jj}^N\Sigma_N\V_N^{(j)}\right| & \leq N_{\max}\mu_1\sqrt{m}(p_e+\delta_2).
\end{align*}
\end{lem}
\proof
Let a Bernoulli random variable $\delta_{ij}$ characterize the event that the
$(i,j)$ entry in $\R$ is observed and is in error. Therefore, $\delta_{ij}=1$ w.p. $pp_e$ and $0$ otherwise. 
Also, let us define $Z_i = \frac{1}{p}\delta_{ij}  u_i^{(t)} N_{\max}$ and $Z = \sum_{i=1}^m~Z_i$. Then, 
\begin{align*}
\mathbb{E}[Z] = \mathbb{E}\left[\sum_{i=1}^{m}~Z_i\right]&= p_e\sum_{i=1}^m~ u_i^{(t)} N_{\max} 
 \overset{\zeta_1}{\leq} N_{\max}p_e\mu_1\sqrt{m},
\end{align*}
where $\zeta_1$ follows from the incoherence of $u_i^{(t)}$. Moreover,
\begin{eqnarray*}
 \text{Var}(Z) = \frac{p_e}{p}(1-pp_e)N_{\max}^2\sum_{i=1}^m|u_i^{(t)}|^2 
 \leq N_{\max}^2\frac{p_e}{p}(1-pp_e) \leq N_{\max}^2\frac{p_e}{p},
\end{eqnarray*}
and $\max_i~|Z_i|  = \frac{1}{p} \max_i ~|u_i^{(t)}N_{ij}| \leq \frac{\mu_1N_{\max}}{p\sqrt{m}}$. Using Bernstein's inequality, we have
\begin{equation*}
\begin{split}
& Pr\left( Z-\mathbb{E}[Z] > N_{\max}\mu_1\sqrt{m}\delta_2 \right) \\
& \qquad \leq
\exp\left( - \frac{N_{\max}^2 \mu_1^2m\delta_2^2/2}{N_{\max}^2\frac{p_e}{p}+\frac{N_{\max}\mu_1}{3p\sqrt{m}}N_{\max}\mu_1\sqrt{m}\delta_2}\right)
\\
& \qquad = \exp\left( - \frac{p \mu_1^2 m \delta_2^2/2}{p_e+\frac{\mu_1^2 \delta_2}{3}} \right) 
\overset{\zeta_2}{\leq} \exp\left( - 3\log n \right)  = \frac{1}{n^3},
\end{split}
\end{equation*}
where 
$\zeta_2$ follows by using the condition from Theorem ~\ref{theo:main_result} that $p>\frac{6\log n}{\mu_1^2m\delta_2^2}(p_e+\frac{\mu_1^2\delta_2}{3})$. 

\noindent Therefore, using the definition from (\ref{eqn:vj_expan}), with probability greater than 
$1-1/n^3$ it holds that 
\begin{eqnarray*}
\left|\frac{1}{B_{jj}}C_{jj}^N\Sigma_N\V_N^{(j)}\right|  =  \left|\frac{1}{p}\sum\limits_{i|ij \in\Omega}u_i^{(t)} [\U_N^{(i)}]^T \Sigma_N\V_N^{(j)}\right| 
=\left|\frac{1}{p}\sum\limits_{i|ij \in\Omega}u_i^{(t)} N_{ij}\right| 
 \leq N_{\max}\mu_1\sqrt{m}(p_e+\delta_2).
\end{eqnarray*}
\endproof
\proof[Proof of Lemma~\ref{lemma:incoherence}]
We bound the largest magnitude of the entries of $\hat{\v}^{(t+1)}$ as follows. 
For every $j\in[n]$, using (\ref{eqn:vj_expan}) we have
\begin{equation}\label{eqn:v_numerator}
\begin{split}
& 2m \left| \hat{v}_j^{(t+1)}\right|  \leq  | \sigma^{\s}\langle \u^{(t)},\u^{\s}\rangle v_j^{\s}|  
\\
&\quad + \left| \frac{\sigma^{\s}}{B_{jj}}(\langle \u^{(t)},\u^{\s}\rangle B_{jj}-C_{jj})v_j^{\s} \right| 
+  \left|   \frac{1}{B_{jj}}C_{jj}^N\Sigma_N \V_N^{(j)}   \right| 
\\
&\quad \overset{\zeta_1}{\leq}   \sigma^{\s}\langle \u^{(t)},\u^{\s}\rangle \frac{\mu}{\sqrt{n}}  + \sigma^{\s}\langle \u^{(t)},\u^{\s}\rangle \frac{\mu}{\sqrt{n}} 
\\
& \quad + \sigma^{\s}(\langle \u^{(t)},\u^{\s}\rangle +\delta_2)\frac{\mu}{\sqrt{n}} 
+ N_{\max}\mu_1\sqrt{m}(p_e+\delta_2)
\\  
& \quad \overset{\zeta_2}{\leq} \sigma^{\s}(3+\delta_2)\frac{\mu}{\sqrt{n}} + N_{\max}\sigma^{\s}\frac{\mu_1}{\sqrt{n}}(p_e+\delta_2) 
 \\
& \quad \overset{\zeta_3}{\leq} \sigma^{\s}\frac{\mu}{\sqrt{n}} (3+\delta_2+8N_{\max}(p_e+\delta_2)) 
\end{split}
\end{equation}
where $\zeta_1$ follows from the fact that $|C_{jj}|\leq (|\langle \u^{(t)},\u^{\s}\rangle| + \delta_2)$ (Lemma C.3,\cite{praneeth}), and Lemma~\ref{lemma:noise_entry_bound},  $\zeta_2$ follows from $\langle \u^{(t)},\u^{\s} \rangle \leq 1$,  and $\zeta_3$ follows since $\mu_1 = 8\mu$.

Furthermore, from Lemma \ref{lemma:vnorm_bound} and using Lemma \ref{lemma:F_bound} and Lemma~\ref{lemma:noise_term_bound}, we have 
\begin{equation}\label{eqn:v_denominator}
\begin{split}
& 2m\|\hat{\v}^{(t+1)}\|  \\
& \qquad \geq  
\sigma^{\s} \langle \u^{(t)},\u^{\s} \rangle - \sigma^{\s}\delta_2\sqrt{1-(\langle \u^{(t)},\u^{\s} \rangle)^2} - \|N_{\text{res}}\|_2  
\\
&\qquad \overset{\zeta_1}{\geq} \sigma^{\s} \langle \u^0,\u^{\s} \rangle   - \sigma^{\s}\delta_2\sqrt{1-(\langle \u^0,\u^{\s} \rangle)^2} -  2N_{\max}\mu_1 p_e \sqrt{mn}
\\
&\qquad \overset{\zeta_2}{\geq} \sigma^{\s} \left(\frac{\sqrt{3}}{2} -\frac{\delta_2}{2} -2N_{\max}\mu_1 p_e \right),  
\end{split}
\end{equation}
where $\zeta_1$ follows from $\dist(\u^{(t)},\u^{\s})\leq \dist(\u^0,\u^{\s}) $ and Lemma \ref{lemma:noise_term_bound}, $\zeta_2$ follows from $\dist(\u^0,\u^{\s}) \leq 1/2$ and  
$\sigma^{\s} = \sqrt{mn}$. 


\noindent Using the two inequalities from (\ref{eqn:v_numerator}) and (\ref{eqn:v_denominator}), we have 
\begin{align*}
&\|\v^{(t+1)}\|_{\infty} = \frac{\|\hat{\v}^{(t+1)}\|_{\infty}}{\|\hat{\v}^{(t+1)}\|_{2}} = 
\frac{\sigma^{\s}\frac{\mu}{\sqrt{n}} (3+\delta_2+8N_{\max}(p_e+\delta_2))}
{\sigma^{\s} \left(\frac{\sqrt{3}}{2} -\frac{\delta_2}{2} -2N_{\max}\mu_1 p_e \right)}.  
\end{align*}
From the condition on $\delta_2$ as specified in  Theorem~\ref{theo:main_result}, 
and by setting $\mu_1=8\mu$, we simplify the above equation as
\begin{align*}
\|\v^{(t+1)}\|_{\infty} \leq \frac{8\mu}{\sqrt{n}} = \frac{\mu_1}{\sqrt{n}}.
\end{align*}
\endproof

\subsection{MEC proofs}
\begin{lem}\label{lemma:MEC_result}
Let $\M,\N,\R$ and $\Omega$ be defined as in Theorem~\ref{theo:main_result}. Let $\hat{\M}^{(T)}$ 
denote the estimate of matrix $\M$ after $T$ iterations of Algorithm~\ref{algo:sigm_altmin}; furthermore, 
let $\tilde{\M}^{(T)} = \text{sign}(\hat{\M}^{(T)})$. Then 
$\tilde{\M}^{(T)}$ satisfies 
\begin{align*}
\frac{1}{\sqrt{mn}}\|P_{\Omega}(\R-\tilde{\M}^{(T)})\|_0 & \leq ~\|P_{\Omega}(\N)\|_F + \|P_{\Omega}(\M - \hat{\M}^{(T)})\|_F. 
\end{align*}
\end{lem}
\proof
Clearly, $\forall i,j \in \Omega \subseteq [m] \times [n]$ it holds that
\begin{align*}
R_{ij}  & \begin{cases}
=\tilde{M}_{ij}^{(T)} & ~~\text{if} ~ |R_{ij} - \hat{M}_{ij}^{(T)}| \leq 1, \\
\neq \tilde{M}_{ij}^{(T)} & ~~\text{otherwise}.
\end{cases} 
\end{align*}
Now, 
$\|P_{\Omega}(\R-\tilde{\M}^{(T)})\|_0$ denotes the number of non-zero entries among the observed entries of the difference matrix $\R-\tilde{\M}^{(T)}$. In other words, 
\begin{align*}
\|P_{\Omega}(\R-\tilde{\M}^{(T)})\|_0 & = \left|  \left\{  i, j\in \Omega  ~\text{s.t.}~ R_{ij} \neq \tilde{M}_{ij}^{(T)} \right\}\right|
 \\
& \leq \left|  \left\{  i,j \in \Omega ~\text{s.t.}~  |R_{ij} - \hat{M}_{ij}^{(T)} | >1  \right\}\right| 
\\
&\leq \sum\limits_{ij\in\Omega} \left| R_{ij} - \hat{M}_{ij}^{(T)}  \right|,
\end{align*}
where the last quantity is the entry-wise $\ell_1$-norm of the matrix $P_{\Omega}(\R-\hat{M}^{(T)})$, denoted by 
$\|P_{\Omega}(\R-\hat{M}^{(T)}\|_1$. Therefore, 
\begin{equation*}
\begin{split}
&\frac{1}{\sqrt{mn}}\|P_{\Omega}(\R-\tilde{\M}^{(T)})\|_0  \\
&\qquad \leq \frac{1}{\sqrt{mn}}\|P_{\Omega}(\R-\M + \M - \hat{\M}^{(T)})\|_1  
\\
& \qquad \leq \frac{1}{\sqrt{mn}}\left(\|P_{\Omega}(\R-\M)\|_1 + \|P_{\Omega}(\M - \hat{\M}^{(T)})\|_1\right) 
\\
& \qquad \leq \|P_{\Omega}(\N)\|_F + \|P_{\Omega}(\M - \hat{\M}^{(T)})\|_F. 
\end{split}
\end{equation*}
\endproof
\proof[Proof of Theorem~\ref{theo:sec_result}]
\begin{small}
\begin{equation*}
\begin{split}
&\text{dist}(\v^{(t+1)},\v^{\s})
\\
&\quad = \|\mathbb{P}_{\v^{\s}_{\perp}}(\hat{\v}^{t+1})\|_2/\|\hat{\v}^{(t+1)}\|_2 
\\
& \quad = \frac{1}{2m} \|\mathbb{P}_{\v^{\s}_{\perp}} (\sigma^\s \langle \u^{(t)},\u^\s \rangle \v^\s -\F + N_{\text{res}} )\|_2/ \|\hat{\v}^{(t+1)}\|_2 
\\
& \quad\overset{\zeta_1}{=}\frac{1}{2m} \|\mathbb{P}_{\v^{\s}_{\perp}} (-\F + N_{\text{res}}) \|_2 / \|\hat{\v}^{(t+1)}\|_2 
\\
& \quad\overset{\zeta_2}{\leq}\frac{1}{2m} \| \F + N_{\text{res}} \|_2 /
\|\hat{\v}^{(t+1)}\|_2 
\\
& \quad\overset{\zeta_3}{\leq} \frac{1}{2m} \left( \|\F\|_2 + \|N_{\text{res}}\|_2 \right) /\|\hat{\v}^{(t+1)}\|_2 
\\
& \quad \overset{\zeta_4}{\leq} \frac{\sigma^\s \delta_2 \sqrt{1- (\langle  \u^{(t)},\u^\s \rangle )^2} + 2N_{\max}p_e \mu_1 \sqrt{mn} }{\sigma^\s \sqrt{1-\text{dist}^2 (\u^{(t)},\u^\s)} - \delta_2 \text{dist}(\u^{(t)},\u^\s) - 2N_{\max}p_e \mu_1 \sqrt{mn}} 
\\
& \quad\overset{\zeta_5}{\leq} \frac{\sigma^\s \left(\delta_2 \text{dist}(\u^{(t)},\u^\s) +  2N_{\max}p_e \mu_1  \right)}{\sigma^\s \left( \frac{\sqrt{3}}{2} - \frac{\delta_2}{2}  -  2N_{\max}p_e \mu_1 \right)} 
\\
& \quad\overset{\zeta_6}{\leq} \frac{1}{4}\text{dist}(\u^{(t)},\u^\s) + \frac{\mu_1 p_e}{\delta_2},
\end{split}
\end{equation*}
\end{small}
where $\zeta_1$ follows since the first term in the numerator is orthogonal to $\v^\s$, $\zeta_2$ follows since $\forall \x,\y, \|\mathbb{P}_{\y}(\x)\|_2\leq \|\x\|_2$, $\zeta_3$ follows from the triangle inequality, $\zeta_4$ follows by using Lemma~\ref{lemma:F_bound} and Lemma~\ref{lemma:noise_term_bound}, $\zeta_5$ follows from the fact that $\sigma^\s = \sqrt{mn}$ and $\text{dist}(\u^{(t)},\u^\s) \leq \text{dist}(\u^0,\u^\s) \leq 1/2$, and finally $\zeta_6$ follows by using the condition on $\delta_2$ from Theorem~\ref{theo:main_result}. 
Using similar arguments, we show that  $\text{dist}(\u^{(t+1)},\u^{\s}) \leq \frac{1}{4}\text{dist}(\v^{(t+1)},\v^\s) + \frac{\mu_1 p_e}{\delta_2}$.
\endproof

\proof[Proof of Theorem~\ref{theo:main_result}]
In order to prove this theorem, firstly we need to bound the error between the true matrix $\M$ and the output of Algorithm~\ref{algo:sigm_altmin} prior to the rounding step. Let us denote the latter as the scaled estimate $\hat{\M}^{(T)}$, where we have 
$\tilde{\M}^{(T)} = \text{sign}(\hat{\M}^{(T)})$. 
By using (\ref{eqn:v_expan}), the difference between $\M$ and (appropriately scaled) $\hat{\M}^{(T)}$ is
\begin{align*}
\M-\hat{\M}^{(T)} &= \M - \u^{(T)}\left[ \sigma^\s \langle \u^{(T)},\u^\s \rangle \v^\s - \F + N_{\text{res}}   \right]^T \\
& = \M - \u^{(T)}(\u^{(T)})^T \u^\s \sigma^\s (\v^\s)^T +\u^{(T)} \F^T 
\\
& \quad - \u^{(T)} N_{\text{res}}^T \\
& = \left( \mathbb{I} - \u^{(T)}(\u^{(T)})^T   \right)\M + \u^{(T)} \F^T - \u^{(T)} N_{\text{res}}^T. 
\end{align*}
\noindent Using the fact that $\|\v^\s\|_2=1$, $\|\u^{(T)}\|_2=1$, and using Lemma~\ref{lemma:F_bound}, Lemma~\ref{lemma:noise_term_bound}, and Corollary~\ref{cor:sec_result}, we have
\begin{equation*}
\begin{split}
& \|\M - \hat{\M}^{(T)}\|_F  \\
& \quad \leq \|\left( \mathbb{I} - \u^{(T)}(\u^{(T)})^T \right)\u^\s \sigma^\s\|_2 + \|\F\|_2 + \|N_{\text{res}}\|_2 
\\
& \quad \leq \sigma^\s \text{dist}(\u^{(T)},\u^\s)  +  \sigma^\s \delta_2  \text{dist}(\u^{(T)},\u^\s) +  2\sigma^\s N_{\max}p_e\mu_1 
\\
& \quad  = \sigma^\s(1+\delta_2) \left(  \frac{\epsilon}{2\|\M\|_F} + \frac{4\mu_1 p_e}{3\delta_2} \right) + 2\sigma^\s N_{\max}p_e\mu_1
\\
& \quad \leq  \frac{\epsilon(1+\delta_2)}{2} + 2\sigma^\s p_e\mu_1 \left(\frac{(3N_{\max}+2)\delta_2+2}{3\delta_2} \right).
\end{split}
\end{equation*}
The theorem then follows by setting $\epsilon^\p = \frac{\epsilon(1+\delta_2)}{2}$ and substituting the value of $\mu_1$.
\endproof
\proof[Proof of Corollary ~\ref{cor:mec}]
Using Lemma~\ref{lemma:MEC_result} and Theorem ~\ref{theo:main_result}, and noting that $\sigma^\s = \sqrt{mn}$, 
the normalized minimum error correction score 
$\tilde{\text{MEC}} = \frac{1}{mn}\|P_{\Omega}(\R-\tilde{\M}^{(T)})\|_0$ can be bounded as
\begin{align*}
\tilde{\text{MEC}} & \leq \frac{1}{\sqrt{mn}} \left( \|P_{\Omega}(\M-\hat{\M}^{(T)})\|_F  + \|P_{\Omega}(\N)\|_F \right) \\
&\leq \epsilon^\p + \frac{2 p_e \mu_1 \sigma^\s}{3\delta_2\sqrt{mn}} \left(2+(2+3N_{\max})\delta_2 \right) + \frac{1}{\sqrt{mn}}\|P_{\Omega}(\N)\|_F 
\\ 
& = \epsilon^\p + \frac{2 p_e \mu_1}{3\delta_2} \left(2+(2+3N_{\max})\delta_2 \right)  + \frac{1}{\sqrt{mn}}\|P_{\Omega}(\N)\|_F. 
\end{align*}
\endproof
\subsection{Characterizing the noise matrix}
\proof[Proof of Lemma~\ref{lemma:noise}]
Noise matrix $\N$ clearly follows the worst case noise model as described in \cite{keshavan} since $\forall i,j \in [m] \times [n], |N_{ij}| \leq N_{\max}$. Let $\Omega^{\p} \subseteq \Omega$ be the set of indices where a sequencing error has occurred, i.e., $\forall (i,j)\in \Omega^{\p}, N_{ij}\neq 0$. 
 Define $\delta_{ij}$ to be a random variable indicating the membership of the index $(i,j)$ in $\Omega^\p$, i.e., $\delta_{ij}=1$ if $(i,j) \in \Omega^\p$, $0$ otherwise. 
 Since sampling and error occur independently, the probability that $\delta_{ij}=1$ is $pp_e$. Therefore, $|\Omega^\p|\approx \mathbb{E}\left[  \sum_{ij}\delta_{ij}\right] = mnpp_e$ w.h.p. 
Using Theorem~\ref{theo:noise} below, we conclude that  
\begin{align*}
\frac{\|P_{\Omega}(\N)\|_2}{p} &\leq 2\frac{|\Omega^\p|\sqrt{m}}{pm\sqrt{n}}N_{\max} 
 = 2N_{\max}p_e\sqrt{mn}. 
\end{align*}
\endproof

\begin{thm}[\cite{keshavan}]\label{theo:noise}
If $\N\in \mathbb{R}^{m \times n}$ ($m \leq n$) is a matrix with entries chosen from the worst case model, i.e., $|N_{ij}| \leq N_{\max}~\forall (i,j)$ for some constant $N_{\max}$, then for a sample set $\Omega$ drawn uniformly at random, it holds that
\begin{align*}
\|P_{\Omega}(\N)\|_2 &\leq \frac{2|\Omega|\sqrt{m}}{m\sqrt{n}}N_{\max}.
\end{align*}
\end{thm}

%

\section*{Acknowledgment}
This work was funded in part by the NSF grant CCF 1320273.

\ifCLASSOPTIONcaptionsoff
  \newpage
\fi



\bibliographystyle{IEEEtran}
\bibliography{ms}
\end{document}